%% file: main.tex
\documentclass{article} %
\usepackage{iclr2027/iclr2027_conference,times}

\input{iclr2027/math_commands.tex}

\usepackage[hidelinks]{hyperref}
\usepackage{url}

\usepackage{graphicx}
\usepackage[inline]{enumitem}
\usepackage{booktabs}
\usepackage[nameinlink,capitalize,noabbrev]{cleveref}

\usepackage{url}
\usepackage[figuresleft]{rotating}
\usepackage{placeins}
\usepackage{ifthen}
\usepackage{stfloats}
\usepackage{siunitx}
\newcolumntype{L}{>{\scriptsize}l}
\newcolumntype{K}{>{\bfseries}S[table-format=1.3]}
\newcolumntype{P}[1]{>{\raggedright\let\newline\\\arraybackslash\hspace{0pt}}m{#1}}

\newtheorem{lemma}{Lemma}

\input{definitions.tex}

\title{Beyond Static Graph World Models:\\ Learning Stochastic Latent Dynamics over Evolving Topologies}

\author{Alex Schutz, Nick Hawes \& Victor-Alexandru Darvariu\\
Oxford Robotics Institute\\
University of Oxford \\
\texttt{\{alexschutz,nickh,victord\}@robots.ox.ac.uk} \\
}

\iclrfinalcopy %
\begin{document}

\maketitle

\begin{abstract}
Graph-based world models have recently emerged as a means of learning transitions over relational state representations. 
However, existing approaches are largely limited to fixed-topology graphs or deterministic, fully observable environments. 
We propose the Graph Dynamics Model (GDM), a world model for graph-structured observations that is designed to handle the more general setting of evolving topologies in stochastic and partially observable environments. 
The GDM uses a sparse recurrent adjacency matrix to model topology updates and perform message passing, together with a recurrent state-space architecture for modelling stochastic transitions. 
Furthermore, we identify a gap in the evaluation of graph-based world models, as existing methods do not provide a means of comparing predicted and true distributions over the joint graph state comprising the interdependent topology, node features, and graph features.  
We therefore introduce the Graph Distribution Distance (GDD) metric, which uses maximum mean discrepancy with a graph kernel to comprehensively compare joint next-state distributions.
We evaluate the GDM across several environments, including stochastic and partially observable settings. We demonstrate that GDM outperforms baseline models and displays zero-shot generalisation on large graphs. 

\end{abstract}

\section{Introduction}
\input{sections/intro}

\section{Related Work}
\input{sections/rel_work}

\section{Background}
\input{sections/background}

\section{Graph Dynamics Model}
\input{sections/method}

\section{Graph Distribution Distance}
\input{sections/evaluation}

\section{Experiments}

\input{sections/results}

\section{Limitations and Future Work}
\input{sections/limitations}

\section{Conclusion}
\input{sections/conclusion}

\subsection*{AI use statement}

In this work, we used generative AI tools to 
provide critical ingredients for proving mathematical claims, 
assist in the writing of proofs,
design or provide feedback on research methodology or experiments, 
and
implement methods.
We have not used generative AI tools to 
formulate mathematical claims,
help develop theoretical models or conceptual frameworks,
propose or refine hypotheses,
interpret results, 
generate synthetic data sets,
or
clean and reformat datasets.
Assisting with translation and
supporting qualitative and thematic data analysis
are not applicable to this work.
Additionally, we used generative AI tools to 
create or edit software code,
summarize or analyse existing literature,
brainstorm,
and
identify relevant literature.
We have reviewed all AI-assisted work. 
All LLM-generated code was verified and tested for correctness.
All mathematical claims and proofs have been reviewed by a qualified researcher.
Methodological feedback was verified by a manual literature review.
We take responsibility for the final content of this work,
including text, claims or artifacts produced with the aid of generative AI.

\subsection*{Reproducibility statement}

All code and data required to reproduce the results of this paper will be uploaded anonymously as supplementary materials and will be released publicly upon publication.
GNN implementations use PyTorch Geometric \citep{fey2019fast}.
Models were trained on a cluster using GPUs with at least 16GB of memory.

\bibliography{references}
\bibliographystyle{iclr2027/iclr2027_conference}

\clearpage
\appendix
\input{sections/appendix}

\end{document}

%% file: iclr2027/math_commands.tex
\usepackage{amsmath,amsfonts,bm}

\def\eqref#1{equation~\ref{#1}}

\def\1{\bm{1}}

\def\va{{\bm{a}}}

\def\ve{{\bm{e}}}

\def\vg{{\bm{g}}}
\def\vh{{\bm{h}}}

\def\vm{{\bm{m}}}

\def\vz{{\bm{z}}}

\def\mA{{\bm{A}}}

\def\mD{{\bm{D}}}

\def\mK{{\bm{K}}}

\def\mQ{{\bm{Q}}}

\def\mU{{\bm{U}}}

\def\mW{{\bm{W}}}
\def\mX{{\bm{X}}}

\DeclareMathAlphabet{\mathsfit}{\encodingdefault}{\sfdefault}{m}{sl}
\SetMathAlphabet{\mathsfit}{bold}{\encodingdefault}{\sfdefault}{bx}{n}

\newcommand{\KL}{D_{\mathrm{KL}}}

%% file: definitions.tex
\newcommand{\nodes}{V}
\newcommand{\edges}{E}
\newcommand{\graph}{G}
\newcommand{\numnodes}{|\nodes|}

\newcommand{\nf}{x}
\newcommand{\gf}{x}
\newcommand{\hn}{\vh}
\newcommand{\hnprior}{\vh^{-}}
\newcommand{\hnpost}{\vh^{+}}
\newcommand{\hg}{\vh}
\newcommand{\hgprior}{\vh^{-}}
\newcommand{\hgpost}{\vh^{+}}
\newcommand{\zn}{\vz}
\newcommand{\znprior}{\vz^{-}}
\newcommand{\znpost}{\vz^{+}}
\newcommand{\zg}{\vz}
\newcommand{\zgprior}{\vz^{-}}
\newcommand{\zgpost}{\vz^{+}}
\newcommand{\adj}{\mA}
\newcommand{\appradj}{\hat{\mA}}

\newcommand{\states}{S}
\newcommand{\actions}{\mathcal{A}}
\newcommand{\transitions}{T}
\newcommand{\observations}{O}
\newcommand{\observationprob}{\Omega}
\newcommand{\reward}{R}
\newcommand{\horizon}{H}
\newcommand{\initdist}{\Delta s_0}
\newcommand{\actionmask}{\mathfrak{A}}

\newcommand{\continuous}{\mathcal{C}}
\newcommand{\categorical}{\mathcal{\delta}}

\newcommand{\freebits}{\lambda_{\text{FB}}}

\newcommand{\loss}{\mathcal{L}}

\newcommand{\numsamples}{\kappa}

\newcommand{\decoder}{\mathcal{D}}
\newcommand{\encoder}{\ve}
\newcommand{\emb}{\operatorname{emb}}
\newcommand{\features}{\mathcal{F}}

\newcommand{\pool}{\operatorname{Pool}}

\newcommand{\paragraphsmall}[1]{\textbf{#1.\;}}

%% file: sections/intro.tex
World models have become popular as a means of learning environment dynamics for applications in reinforcement learning (RL) and planning, enabling reduced sample complexity, improved generalisation, and offline learning \citep{haWorldModels2018,hafner2024masteringdiversedomainsworld}. 
Most existing world model frameworks are designed for vector or pixel-based observations, which limits their applicability to environments with rich relational structures, such as graphs.
Graph-based environments, where an agent interacts with a system comprising nodes, edges, and their respective features, are encountered in a number of contexts, including social networks~\citep{pmlr-v139-meirom21a}, navigation~\citep{lu2021mgrl}, and infrastructure~\citep{amorosa2024multi}.
In many of these applications, the topology of the graph is variable, with edges being added or removed as the environment evolves.
An advantage of graph-based models is the potential for zero-shot generalisation to larger environments, as a model trained on smaller graphs can then be applied to larger graphs where direct training is infeasible.

\begin{figure}[tbh]
	\centering
	\includegraphics[width=\linewidth]{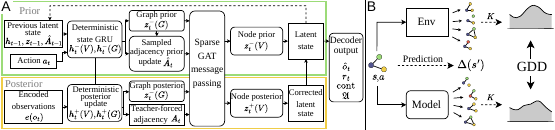}
	\caption{A) Architecture of the Graph Dynamics Model. B) Illustration of the GDD evaluation metric, which compares the predicted and true distributions of the next state.}
	\label{fig:gdm_architecture}
\end{figure}

Despite the prevalence of graph-based environments, graph-based world models are still an emerging area of research.
In environments with evolving topology, modelling becomes more difficult, as topology predictions and message passing are intertwined, and the model must learn to predict accurate topology in order to perform message passing effectively.
As such, current graph-based world models are mostly limited to fixed topology \citep{fengGraphWorldModel2025,chenGraphWorldModel2026}, do not exploit the topology for message passing \citep{karacelebiLearningAdHoc2026}, or are effective only in deterministic fully-observable settings \citep{songUnderstandingRolloutError2026}.
We overcome these limitations by proposing the \textbf{Graph Dynamics Model (GDM)}, a world model for graph-structured observations with evolving topology, applicable to stochastic and partially observable environments.
The GDM uses a sparse recurrent adjacency matrix, inspired by SparseGAT~\citep{yeSparseGraphAttention2019}, with a recurrent state-space architecture \citep{hafnerLearningLatentDynamics2019}, in order to predict stochastic transitions in a latent space (\Cref{fig:gdm_architecture}A).
Topology prediction is one of the most important aspects of a graph-based world model, and the sharp adjacency updates facilitate accurate feature distribution predictions across the joint graph state.

There does not currently exist a standardised protocol for evaluating world models.
In the absence of a downstream task, world models are typically evaluated based on the accuracy of their predictions compared to the environment.
However, this is insufficient for stochastic environments, where the model must learn to capture the full distribution of state predictions rather than simply the mean.
Furthermore, when the state consists of multiple factors, such as a graph environment with different node features, graph features, and topology, the evaluation must consider the full joint state rather than comparing factors independently.
In this work, we propose a novel evaluation metric for next-step prediction of stochastic graph states based on maximum mean discrepancy (MMD) \citep{grettonKernelTwosampleTest2012}.
The evaluation metric, which we call the \textbf{Graph Distribution Distance (GDD)} metric (\Cref{fig:gdm_architecture}B), captures the joint distribution of the graph topology, node features, and graph features, and is zero if and only if the predicted distribution matches the true distribution. 
This provides a principled way to compare predictions of graph-based world models in stochastic environments.

Our contributions can be summarised as follows.
\begin{enumerate*}[nosep]
	\item We propose the Graph Dynamics Model (GDM), a world model for graph-structured observations with evolving topology, applicable to stochastic and partially observable environments.
	\item We introduce a novel evaluation metric, GDD, for measuring world model quality using the full joint next-state distribution of graph topology, node features, and graph features, and validate it using per-factor distribution metrics.
	\item We conduct a study of the proposed model and metric on a set of environments with varying characteristics at different graph sizes not encountered during training, demonstrating the GDM's ability to generalise to unseen graph sizes and outperform state-of-the-art baselines on multi-step rollouts.
	\item We perform an ablation of the proposed model, demonstrating the importance of the recurrent adjacency sampling.
\end{enumerate*}

%% file: sections/rel_work.tex
\paragraphsmall{World Models}
In model-based RL, a world model is a model of the environment's transition function, which predicts the next state given the current state and action, commonly also including a reward and continuation signal prediction.
The RSSM framework \citep{hafnerLearningLatentDynamics2019,hafner2024masteringdiversedomainsworld} models environment dynamics in latent space through recurrent updates of deterministic and stochastic states.
Earlier latent state-space approaches include the Deep Variational Bayes Filter \citep{karlDeepVariationalBayes2017}, based on structured variational inference, and the Probabilistic Recurrent State-Space Model \citep{doerrProbabilisticRecurrentStateSpace2018}, using Gaussian process dynamics.
More recent approaches employ alternative representations for latent dynamics, including autoregressive transformer world models \citep{micheli2024efficientworldmodelscontextaware,agarwal2024learning}, structured state-space sequence models \citep{mattes2024hieros}, and object-centric latent dynamics \citep{mosbach2025sold}.

\paragraphsmall{Graph-Based World Models}
Graph-based world models apply the world model paradigm to graph-structured observations.
Building on the action-node representation of \citet{fengGraphWorldModel2025}, \citet{songUnderstandingRolloutError2026} study graph world models operating in state space, formulating GNN-based world models as 
$\sigma(\tilde{\mA}\mX\mW_1 + \mU a\mathbf{1}^\top)\mW_2$,
with $\tilde{\mA}=\mD^{-1/2}\mA\mD^{-1/2}$,
where $\mX$ is the node feature matrix, $\mU$, $\mW_1$ and $\mW_2$ are learned parameters, and $\mA$ is predicted by a learned  function of the pairwise node feature embeddings.
Direct state space transitions can accurately capture deterministic dynamics, but are less effective in stochastic settings, where the model must capture the full distribution of next states rather than just the mean.
\citet{karacelebiLearningAdHoc2026} propose the G-RSSM model for learning ad-hoc network dynamics, which generalises the latent-space RSSM to graphs by including a recurrent state vector for each node, being applicable to stochastic and partially observable environments.
G-RSSM uses all-pairs message passing to update the latent state, rather than exploiting the underlying topology of the graph, only using a GNN to encode observations of the true state.
Several other works have proposed graph-based world models for specific applications, such as structured origami folding \citep{huangLearn2FoldStructuredOrigami2026}, LEO satellite networks \citep{liuGraphWorldModel2026}, and structural dynamics in graphs \citep{wangStructuralDynamicsGraph2026}. 
These approaches rely on domain-specific knowledge and are not applicable to general graph-based environments.

\paragraphsmall{World Models on Learned Graphs} 
Several works integrate graph structures into world models where the observation space is not a graph, but structure can be inferred.
\citet{fengLearningDynamicAttributefactored2023} build a world model for object interactions, modelling object relationships using dynamic Bayesian networks.
Dynamic Neural Relational Inference \citep{graberDynamicNeuralRelational2020} learns relations for each entity pair at each time step.
In their survey paper, \citet{liuGraphWorldModels2026} highlight that existing graph world models do not adequately address topological plasticity and probabilistic modelling.

\paragraphsmall{World Model Evaluation}
World model evaluation procedures vary depending upon both the environment being modelled (e.g., deterministic versus stochastic) and the downstream task for which the model is intended.
Many works rely on downstream planning performance as a measure of world model quality \citep{hafnerLearningLatentDynamics2019,hafnerMasteringAtariDiscrete2022,yildiz2021continuous,chungThinkerLearningPlan2023a}, which can be computationally expensive.
In the absence of a downstream task, world models are often evaluated using non-distributional metrics such as accuracy and MSE \citep{schiewerExploringLimitsHierarchical2024,hao2025neural}.
The stochastic world model benchmark proposed in \citep{barsainyanSTORIBenchmarkTaxonomy2025} finds that ``world models dramatically underestimate environmental variance''. %
Distributional metrics for world model evaluation are common in video-generative applications \citep{gao2024vista,barNavigationWorldModels2025}, being much less common in model-based RL applications \citep{sedlmeier2021quantifyingmultimodalityworldmodels}.

%% file: sections/background.tex
\paragraphsmall{Graph-based POMDP Formulation}\label{sec:mdp}
Let an attributed graph be $\graph=(\nodes,\edges, \{\nf(v)\}_{v\in\nodes}, \gf(\graph))$, comprising a set of nodes $\nodes$, edges $\edges\subseteq\nodes\times\nodes$, node feature vectors $\nf(v)\in\mathcal X_n$, and a graph feature vector $\gf(\graph)\in\mathcal X_g$.
Consider a tuple $\mathcal{P} = \langle \states, \actions, \transitions, \observations, \observationprob, \reward, \horizon, \initdist \rangle$
defining a partially observable Markov decision process (POMDP) with state space $\states$, action space $\actions$, transition probability function ${\transitions: \states \times \actions \rightarrow \Delta \states}$, observation space $\observations$, observation probability function ${\observationprob: \states \times \actions \rightarrow \Delta \observations}$, reward function ${\reward:\states\times\actions\to\mathbb{R}}$, horizon $\horizon$, and initial state distribution $\initdist \in \Delta \states$.
We say that $\mathcal{P}$ is fully observable if $\states = \observations$ and $\observationprob(o \mid s')=\mathbb{I}[o=s']$, and partially observable otherwise.
When $\observations$ is a set of feature-equipped graphs, i.e., each $o\in \observations$ corresponds to some $\graph$, we call $\mathcal{P}$ a graph-based environment.
We consider dynamic graphs with evolving topology, where the edge set $\edges$ is gradually updated, in contrast to temporal graphs, where the topology is highly dynamic and instantaneous snapshots do not represent the overall structure \citep{skardingFoundationsModelingDynamic2021}.
Here we will assume that $(\nodes,\edges)$ is always fully observable, but the node and graph features may be partially observable. 
We define $\actions$ to be a set of node selections. 
Complex actions such as edge additions or rewiring can be composed from multiple node selections.
For a state $s\in\states$, the action mask returns the set of valid nodes: $\actionmask:\states\to 2^{\nodes}$.
The set of valid actions in state $s$ is then
$
\actions(s)
=
\left\{
a\subseteq \actionmask(s)\mid |a|=k
\right\}
$,
for some problem-defined $k$.

\paragraphsmall{Maximum Mean Discrepancy}\label{sec:mmd}
As we have argued, comparing distributions predicted by graph-based world models is essential in stochastic environments. 
However, this has not been studied in prior work. 
We build on the versatile MMD for this purpose, which is a kernel-based measure of the distance between two probability distributions $P$ and $Q$ defined on a common space $\mathcal X$ \citep{grettonKernelTwosampleTest2012}. 
Let $K:\mathcal X\times\mathcal X\to\mathbb R$ be a positive definite kernel, with associated reproducing kernel Hilbert space (RKHS) $\mathcal H_K$. 
Assuming the required expectations are finite, the MMD is defined as:
{\small
\begin{equation}
\operatorname{MMD}_K(P,Q)
=
\sup_{f\in\mathcal H_K:\|f\|_{\mathcal H_K}\leq 1}
\left(
\mathbb E_{x\sim P}[f(x)]
-
\mathbb E_{y\sim Q}[f(y)]
\right).
\end{equation}
}%
Each distribution can be represented in $\mathcal H_K$ by its kernel mean embedding:
$
\mu_P=\mathbb E_{x\sim P}[K(x,\cdot)]
$
and
$
\mu_Q=\mathbb E_{y\sim Q}[K(y,\cdot)],
$
in which case
$
\operatorname{MMD}_K(P,Q)
=
\|\mu_P-\mu_Q\|_{\mathcal H_K}
$.
Using the reproducing property, its squared value can equivalently be written as:
{\small
\[
\operatorname{MMD}_K^2(P,Q)
=
\mathbb E_{x,x'\sim P}[K(x,x')]
+
\mathbb E_{y,y'\sim Q}[K(y,y')]
-
2\mathbb E_{\substack{x\sim P\\y\sim Q}}[K(x,y)],
\]
}%
where $x,x'$ are independent draws from $P$, $y,y'$ are independent draws from $Q$, and the draws from $P$ and $Q$ are mutually independent.
For i.i.d. samples ${X=\{x_i\}_{i=1}^n\sim P^n}$ and ${Y=\{y_j\}_{j=1}^m\sim Q^m}$, the biased empirical estimator of the squared MMD is
{\small
\begin{equation}\label{eqn:mmd_estimate}
\widehat{\operatorname{MMD}}_K^2(X,Y)
=
\frac{1}{n^2}\sum_{i=1}^n\sum_{i'=1}^n K(x_i,x_{i'})
+
\frac{1}{m^2}\sum_{j=1}^m\sum_{j'=1}^m K(y_j,y_{j'})
-
\frac{2}{nm}\sum_{i=1}^n\sum_{j=1}^m K(x_i,y_j).
\end{equation}
}%
A kernel $K$ is characteristic if its kernel mean embedding
$P\mapsto \mu_P$
is injective over the class of distributions under consideration, meaning that $\mu_P=\mu_Q$ implies $P=Q$. 
Thus, when $K$ is characteristic,
$
\operatorname{MMD}_K(P,Q)=0
\Leftrightarrow
P=Q
$
\citep{sriperumbudurHilbertSpaceEmbeddings2010}.

\paragraphsmall{Recurrent State-Space Models}
The recurrent state-space model (RSSM) framework \citep{hafnerLearningLatentDynamics2019} represents the environment's state in a latent space, composed of a \textit{deterministic state} $\vh_t$ and a \textit{stochastic state} $\vz_t$.
In a forward pass, also called \textit{imagination}, the deterministic state is updated recurrently, while the stochastic state is sampled from a learned \textit{prior distribution} $p(\vz_t \mid \vh_t)$.
During training, the \textit{posterior update} $q(\vz_t \mid \vh_t, o_t)$ is used to correct the latent state based on the observation $o_t$, and the model is trained to minimise the KL divergence between $p$ and $q$, as well as the reconstruction loss between the decoded latent state $\decoder(\vh_t, \vz_t)$ and $o_t$.

%% file: sections/method.tex
\subsection{Architecture}

The GDM uses an RSSM-style architecture designed for dynamic graph-structured observations, as shown in \Cref{fig:gdm_architecture}A.
Rather than relying on a single global latent state, the model assigns a latent vector $\hn(v)$ and $\zn(v)$ for each node $v \in \nodes$, as well as a latent vector $\hg(\graph)$ and $\zg(\graph)$ for the graph as a whole.
Key to the model is the sampled recurrent adjacency matrix $\appradj_t$, enabling message passing on the predicted topology and including it in the state, rather than only reconstructing it in the decoder as per G-RSSM \citep{karacelebiLearningAdHoc2026}.
While the stochastic components of the latent state $\vz$ and $\appradj$ are factorised, they are not independent, and the model captures their dependencies through conditional distributions.
During imagination, the model predicts states using the prior dynamics, denoted~$^-$, while during training, the model uses the posterior dynamics, denoted~$^+$.
Hereafter, the latent states $\vh$ and $\vz$ refer to the posterior state during posterior inference, and the prior state otherwise.

\paragraphsmall{Encoder}
The encoder produces embeddings from the observation $o$, factorised into node embeddings $\encoder_v(o)$ for each node $v$ and a graph embedding $\encoder_g(o)$.
A feature embedding is produced by passing each feature through a feature-specific learned linear transformation and taking the mean of the resulting embeddings.

\paragraphsmall{Recurrent Dynamics and Posterior-Informed Deterministic State}
For node $v$, let $\va_t(v)$ be its action embedding, produced by a linear transformation of the action set.
The deterministic prior states are then updated recurrently:
{\small
\begin{align*}
\hnprior_{t}(v)
&=
\operatorname{LayerNorm}
\left(
\operatorname{GRU}_n
\left(
\zn_{t-1}(v),\va_{t}(v)
\mid
\hn_{t-1}(v)
\right)
\right),\\
\hgprior_t(\graph)
&=
\operatorname{LayerNorm}
\left(
\operatorname{GRU}_g
\left(
\zg_{t-1}(\graph), \pool_{v \in \nodes}(\va_t(v)) 
\mid
\hg_{t-1}(\graph)
\right)
\right).
\end{align*}
}%
In an ordinary RSSM model, the observation is used to update only the stochastic component of the latent state, while the deterministic component is updated solely based on the previous state and the current input.
While this model is effective for pixel-based observations, where there can be much redundancy in the observation, we found it to be less effective for dense observation features.
As such, during training, the deterministic states are updated: 
$
\hnpost_t
=
\hnprior_t
+
\operatorname{sigmoid}
\left(
\vg_t
\right)
\odot
\Delta\vh_t
$, 
where
\[
\resizebox{\linewidth}{!}{%
  $\displaystyle
\Delta\vh_t(v) \| 
\vg_t(v)
=
\operatorname{MLP}^{+}_{n}
\left(
\hnprior_{t}(v),\ve_v(o_t),\ve_g(o_t)
\right),
\quad
\Delta\vh_t(\graph) \| 
\vg_t(\graph)
=
\operatorname{MLP}^{+}_{g}
\left(
\hgprior_t(\graph),\ve_g(o_t),\pool_{v \in \nodes}(\ve_v(o_t))
\right).
  $
}
\]

\paragraphsmall{Adjacency Prediction and Message Passing}
The graph stochastic state is sampled from a categorical distribution conditioned on the graph latent state:
{\small
\begin{align*}
	&\zgprior_t(\graph) \sim p_g(\zg_t(\graph) \mid \hgprior_t(\graph), \pool_{v \in \nodes}(\hnprior_t(v))),
	&\zgpost_t(\graph) \sim q_g(\zg_t(\graph) \mid \hgpost_t(\graph), \pool_{v \in \nodes}(\hnpost_t(v)), \encoder_g(o_t)).
\end{align*}
}%
We apply Unimix of 1\% to all categorical sampling as per \citet{hafner2024masteringdiversedomainsworld}.
The adjacency matrix $\appradj_t$ is predicted using an update network, predicting edge additions and removals under the assumption of incremental changes.
For each ordered node pair $(i,j)$, we calculate the edge update logit
$
{\ell^{\Delta\text{adj}}_{i j}
=
\operatorname{MLP}_{\text{adj}}\!\left(
\hnprior_{t}(i), \hnprior_{t}(j), \hgprior_{t}(\graph), \zgprior_{t}(\graph), \appradj_{t-1}(i,j)
 \right)}
$.
We then apply Hard Concrete sampling \citep{louizos2018learning} to obtain a differentiable sample:
{\small
\begin{align*}
\tilde{\ell}_{i j}
=
\ell^{\Delta\text{adj}}_{i j}
+
\beta\log\!\left(\frac{-\gamma}{\zeta}\right),
&&
s_{ij}
=
\operatorname{sigmoid}\!\left(
\frac{
\log u_{ij}-\log(1-u_{ij})+\tilde{\ell}_{ij}
}{\beta}
\right),
&&
u_{ij}\sim\operatorname{U}(0,1),
\end{align*}
}%
with Hard Concrete gates given by $c_{ij}
=
\operatorname{clip}_{[0,1]}
\left(
s_{ij}(\zeta-\gamma)+\gamma
\right)
$.
The adjacency matrix is updated as
$
\appradj_{t}(i,j)
=
\appradj_{t-1}(i,j) 
\mathbin{\operatorname{xor}} 
\mathbf{1}[c_{ij}>0]
$.
The corresponding differentiable next-edge weight is given by
$
b_{ij}
=
(1-\appradj_{t-1}(i,j))c_{ij}
+
\appradj_{t-1}(i,j)(1-c_{ij})
$.
To preserve the hard topology in the forward pass while allowing downstream losses to differentiate through every edge update, we define the straight-through next-edge weight 
$
\bar b_{ij}
=
b_{ij}
+
\operatorname{sg}\!\left(
\appradj_t(i,j)-b_{ij}
\right)
$, 
where $\operatorname{sg}$ denotes stop-gradient.
We then perform SparseGAT-style message passing on $\appradj_{t}$, with normalised weights derived from $\bar{b}_{ij}$ being shared across all attention heads and layers per \citet{yeSparseGraphAttention2019}.
We inject $[\hg_t(\graph), \zg_t(\graph)]$ into the first layer, and apply a learned root connection followed by layer normalisation:
{\small
\[
\vm^{(\ell+1)}_t
=
\operatorname{LayerNorm}
\left(
\mW_{\mathrm{root}}^{(\ell)}\vm^{(\ell)}_t
+
\operatorname{SparseGAT}^{(\ell)}
\right),
\qquad \vm_t^{(0)}=\vh_t.
\]
}%
On the posterior path, $\appradj_t$ is teacher-forced to the ground truth adjacency matrix from the observation, with $b_{ij} = \adj_{t}(i,j)$.
Finally, the node stochastic state is sampled from a categorical distribution:
{\small
\begin{align*}
	&\znprior_{t}(v) \sim p_n(\zn_{t}(v) \mid \vm_t^{(L)-}(v), \zgprior_{t}(\graph)),
	&\znpost_{t}(v) \sim q_n(\zn_{t}(v) \mid \vm_t^{(L)+}(v), \zgpost_{t}(\graph), \encoder_v(o_t)).
\end{align*}
}%

\paragraphsmall{Decoder}
The decoder is used to predict the observation from the current latent state.
The decoder is factorised into a mean component $\theta$ and a stochastic innovation component $I$:
{\small
\begin{equation}\label{eqn:decoder}
	\begin{split}
		\decoder_g(\hgprior_t(\graph),\zg_t(\graph))&=\theta(\hgprior_t(\graph))+I\bigl(\hgprior_t(\graph),\zg_t(\graph)-\mathbb{E}_{p_g(\zg_t(\graph)\mid \hgprior_t(\graph))}[\zg_t(\graph)]\bigr), \\
		\decoder_v(\hnprior_t(v),\zn_t(v))&=\theta(\hnprior_t(v))+I\bigl(\hnprior_t(v),\zn_t(v)-\mathbb{E}_{p_n(\zn_t(v)\mid \vm_t^{(L)-}(v), \zgprior_t(\graph))}[\zn_t(v)]\bigr).
	\end{split}
\end{equation}
}%
Here, the prior deterministic state $\vh^{-}$ is always used, ensuring that the decoder does not bypass the stochastic state through the observation-informed $\vh^{+}$.
This formulation provides a clear separation between the deterministic and stochastic components of the model's predictions.
The shared state produced by $\decoder$ is then passed to feature-specific heads to predict the observed node and graph features, and the action mask, reward, and continuation signal.

\subsection{Training}

The GDM is trained to minimise the standard RSSM loss, with an additional $\vh$ alignment term:
\begin{equation}\label{eqn:loss}
\loss = \beta_\text{pred} \loss_\text{pred} + \loss_{\text{KL}} + \beta_h\loss_h.
\end{equation} 
The prediction loss is the average over all feature prediction terms:
{\small
\begin{equation*}
\loss_{\text{pred}}
=
\frac{1}{\numnodes |\features_n| }
\sum_{v\in \nodes} 
\sum_{f\in\features_n}\ell_f^{(n)}
+
\frac{1}{|\features_g| }
\sum_{f\in\features_g}\ell_f^{(g)}
+
\ell_{\text{adj}}
+
\ell_{\text{action}}
+
\ell_{\text{reward}}
+
\ell_{\text{continue}}
,
\end{equation*}
}%
where $\ell_f$ is MSE for scalar features and CE for categorical features, $\ell_{\text{adj}}$ is BCE on $\appradj$, $\ell_{\text{action}}$ is BCE on the action mask, $\ell_{\text{reward}}$ is MSE on the reward, and $\ell_{\text{continue}}$ is BCE on the continuation signal.
The KL loss is the weighted sum of the node KL, graph KL, and adjacency KL:
$
\loss_{\text{KL}}
=
{\beta_n \loss_{\text{KL}}^n
+
\beta_g \loss_{\text{KL}}^g
+
\beta_{\text{adj}}\loss_{\text{KL}}^{\text{adj}}
}
$.
For both $\loss_{\text{KL}}^n$ and $\loss_{\text{KL}}^g$, we separate the loss into a weighted sum of dynamics KL and a representation KL \citep{hafner2024masteringdiversedomainsworld}:
$
\beta_\text{dyn}
\left[
\KL
\left(
\operatorname{sg}(q)\,\|\,p
\right)
\right]_{\text{FB}}
+
\beta_\text{rep}
\left[
\KL
\left(
q\,\|\,\operatorname{sg}(p)
\right)
\right]_{\text{FB}}
$,
where $[x]_{\text{FB}}=\max(x,\freebits)$ is the free-bits floor.
For adjacency, using posterior edge probability $q_{ij}$ (teacher-forced) and prior edge probability $p_{ij}$,
\[
\resizebox{\linewidth}{!}{%
  $\displaystyle
\loss_{\text{KL}}^{\text{adj}}
=
\frac{1}{\numnodes^2}
\sum_{i,j\in \nodes}\left(
q_{ij}\log\frac{q_{ij}}{p_{ij}}
+
(1-q_{ij})\log\frac{1-q_{ij}}{1-p_{ij}}
\right),
\text{ where }
p_{ij}
=
\begin{cases}
\operatorname{sigmoid}\!\left(\ell^{\Delta\text{adj}}_{ij}\right),
&
\appradj_{t-1}(i,j)=0,
\\[2mm]
1-\operatorname{sigmoid}\!\left(\ell^{\Delta\text{adj}}_{ij}\right),
&
\appradj_{t-1}(i,j)=1.
\end{cases}
$
}%
\]
Finally, we encourage the deterministic prior and posterior states to be aligned:
\[
\resizebox{\linewidth}{!}{%
  $\displaystyle
2\loss_h
=
\frac{1}{\numnodes}\sum_{v \in \nodes}
\left(
\left\|
\hnprior(v)
-
\operatorname{sg}(\hnpost(v))
\right\|^2
+ 
\left\|
\hnpost_t(v)-\hnprior_t(v)
\right\|^2
\right)
+
\left\|
\hgprior(\graph)
-
\operatorname{sg}(\hgpost(\graph))
\right\|^2
+
\left\|
\hgpost_t(\graph)-\hgprior_t(\graph)
\right\|^2.
$
}%
\]

%% file: sections/evaluation.tex
We propose using MMD in order to evaluate the quality of a world model in a stochastic environment with joint distributions over topology, node features, and graph features.
A whole-graph metric such as MMD is preferable to evaluating the distribution of individual factors separately, as it captures dependencies between the factors and provides a more complete measure of the model's performance.
Using a characteristic kernel, the MMD is 0 only when the predicted distribution matches the true distribution, so we can determine when a model is truly capturing the full joint dynamics of the transition function.
While previous works have defined graph kernels for use in MMD \citep{borgwardtProteinFunctionPrediction2005,borgwardtGRAPHKERNELSDISEASE2006}, these kernels are not characteristic, and do not consider all feature types. Therefore, we construct a new characteristic kernel for attributed graphs and use it to define GDD.

Suppose that the node features can be partitioned into a set of categorical features indexed by $\features_n^\categorical$ and a set of continuous features indexed by $\features_n^\continuous$, and similarly for the graph features with index sets $\features_g^\categorical$ and $\features_g^\continuous$.
For a finite index set $I$ of continuous features, define
{\small
\[
D^\continuous_I(x,x')=
\frac{1}{|I|}
\sum_{f\in I}
\frac{|x_f-x_f'|^2}{\sigma_f^2},
\qquad \sigma_f > 0,
\]
}%
where $x_f$ is the value of feature $f$ and $\sigma_f$ is a fitted bandwidth.
For continuous node and graph features, we use the multi-bandwidth Radial Basis Function (RBF) kernel:
\[
\resizebox{\linewidth}{!}{%
  $\displaystyle
  K_n^\continuous(G,G')=
  \frac{1}{|\mathcal M|}\sum_{r\in\mathcal M}
  \exp\left(-\frac{
  \sum_{v\in\nodes}D^\continuous_{\features_n^\continuous}(\nf(v),\nf'(v))}{2\numnodes r^2}\right),
  \enspace
   K_g^\continuous(G,G')=
  \frac{1}{|\mathcal M|}\sum_{r\in\mathcal M}
  \exp\left(-\frac{D^\continuous_{\features_g^\continuous}(\gf(G),\gf(G'))}{2r^2}\right)
  $
}
\]
where $0 < r < \infty$ for every $r$ in the bandwidth multiplier set $\mathcal M$.
For a finite index set $I$ of categorical features, we use the weighted Hamming distance
{\small
\[
D^\categorical_I(x,x')=
\frac{1}{|I|}
\sum_{f\in I}
\lambda_f\mathbf 1[x_f\ne x_f'],
\qquad \lambda_f>0.
\]
}%
The node and graph categorical kernels are thus defined:
{\small
\[
K_n^\categorical(G,G')=
\exp\left(-\frac{1}{\numnodes}
\sum_{v\in\nodes}
D_{\features_n^\categorical}^\categorical(\nf(v),\nf'(v))\right),
\qquad
K_g^\categorical(G,G')=
\exp\left(-D_{\features_g^\categorical}^\categorical(\gf(G),\gf(G'))\right).
\]
}%
Where no features are present for a given type, we define $D_{\features}(G,G')=0$.
For the topology, we use the Jaccard kernel
{\small
\[
K_J(G,G')=\frac{|\edges_G\cap\edges_{G'}|}
{|\edges_G\cup\edges_{G'}|},
\quad
\text{ where } K_J(\emptyset,E)=0 \text{ when } E\ne\emptyset 
\text{ and } K_J(\emptyset,\emptyset)=1.
\]
}%
Finally, the joint kernel is defined as:
{\small
\begin{equation}\label{eqn:joint_kernel}
K_\mathrm{joint}(G,G') = K_J(G,G') K_n^\continuous(G,G') K_n^\categorical(G,G') K_g^\continuous(G,G') K_g^\categorical(G,G').
\end{equation}
}

\begin{lemma}\label{lemma:characteristic_joint_kernel}
	Assume that all graphs have the same finite, aligned node set. 
	Assume also that the node and graph feature index sets are finite and that every categorical feature has a finite value space.
	Then $K_\mathrm{joint}$ is a characteristic kernel on this space of attributed graphs.
	\textup{(Proof in \Cref{app:proof_characteristic_joint_kernel}).}
\end{lemma}

As the joint kernel is characteristic, the kernel can distinguish between different probability distributions over the space of attributed graphs.
However, in the joint kernel formulation, any arbitrarily poor factor can drive the entire kernel toward $0$,
causing $\operatorname{MMD}{K_{\mathrm{joint}}}$ to be dominated by that factor, compressing the range of the metric.
As such, in order to have a useful metric for distinguishing both good and bad predictions, we are motivated to include additive terms which consider the factors independently.
We define individual kernels for continuous and categorical features:
\[
\resizebox{\linewidth}{!}{%
$
\displaystyle
k^\continuous_f(x,x')=\frac{1}{|\mathcal{M}|}\sum_{r\in\mathcal{M}}
\exp\left(-\frac{|x-x'|^2}{2r^2\sigma_f^2}\right),
f \in \features_n^\continuous\cup\features_g^\continuous,
\qquad
k^\categorical_f(x,x')=\exp\left(-\lambda_f\mathbf 1[x\ne x']\right),
f \in \features_n^\categorical\cup\features_g^\categorical.
$
}%
\]
We then define the additive node and graph kernels as
\[
\resizebox{\linewidth}{!}{%
  $\displaystyle
K_\mathrm{add}^n(G,G')=
\frac{1}{\numnodes|\features_n|}
\sum_{v\in\nodes}\sum_{f\in\features_n}
k_f(\nf_{f}(v),{\nf'_{f}(v)}),
\enspace
K^g_\mathrm{add}(G,G')=
\frac{1}{|\features_g|}\sum_{f\in\features_g}
k_f(\gf_f(\graph),{\gf'_f(\graph')}).
$
}
\]
For the additive topology term, we again use $K_J$.
The full additive graph kernel is
{\small
\[
K_\mathrm{add}(G,G')=
w_nK_\mathrm{add}^n(G,G')+
w_gK_\mathrm{add}^g(G,G')+
w_eK_J(G,G'),
\]
}%
where $w_n,w_g,w_e\geq0$ control the relative importance of the three terms.
Finally, the factorised graph kernel is
{\small
\begin{equation}
	K(G,G')=K_\mathrm{add}(G,G')+\epsilon K_\mathrm{joint}(G,G'),
	\qquad \epsilon>0.
\end{equation}
}%

\begin{lemma}\label{lemma:characteristic_kernel}
	Under the assumptions of \Cref{lemma:characteristic_joint_kernel}, $K$ is a characteristic kernel on the space of attributed graphs. \textup{(Proof in \Cref{app:proof_characteristic_kernel}).}
\end{lemma}

Using the factorised graph kernel, we define the GDD metric as the MMD between the predicted and ground truth graph distributions:
{\small
	\begin{equation}\label{eqn:GDD}
	\operatorname{GDD}(P,Q)=\operatorname{MMD}_K(P,Q).
\end{equation}
}%
The corresponding squared GDD decomposes as
{\small
\[
\operatorname{GDD}^2=
w_n\operatorname{MMD}_{K^n_\mathrm{add}}^2+
w_g\operatorname{MMD}_{K^g_\mathrm{add}}^2+
w_e\operatorname{MMD}_{K_J}^2+
\epsilon\operatorname{MMD}_{K_\mathrm{joint}}^2.
\]
}%
Thus the additive terms retain sensitivity to errors in individual factors, while any $\epsilon>0$ preserves distribution distinguishing.
We set $w_n=w_g=w_e=1/3$ and $\epsilon = 1$, providing equal weighting between the additive and joint terms, and between the three additive terms.
In practice we use the biased MMD approximation in \Cref{eqn:mmd_estimate} to calculate the GDD from sampled distributions. 
We demonstrate in \Cref{app:zero-GDD} that in-distribution samples produce a near-zero GDD estimate.

%% file: sections/results.tex
\paragraphsmall{Baselines}
From \citet{songUnderstandingRolloutError2026}, we use the Error-Aware GWM (EA-GWM) as a baseline, which models deterministic transitions directly in the state space, and uses several regularising terms to mitigate multi-step rollout error. 
We also consider a Vanilla GWM baseline, which is EA-GWM without the regularisation terms.
We also include G-RSSM \citep{karacelebiLearningAdHoc2026}, which models stochastic transitions in latent space and uses all-pairs message passing to update the latent state rather than relying on a predicted adjacency matrix.
For all baselines we adapt the encoders and decoders to the environments.
Full adaptation details are provided in \Cref{app:baseline_adaptation}.

\paragraphsmall{Environments}
We consider four evolving-topology environments with different levels of stochasticity and observability.
We defer full environment descriptions to \Cref{app:environment_descriptions}.\\[0.2em]
\textit{Robust Graph Construction (RGC).} Given an initial graph, the agent selects node pairs to construct new edges between, in order to minimise the probability of disconnection under node removals~\citep{darvariuGoaldirectedGraphConstruction2021}. The environment is deterministic and fully observable.\\[0.2em]
\textit{Opinion Dynamics (OD).} The agent spreads a target opinion in an evolving network by selecting a set of nodes to try to activate at each step. At each step, a subset of nodes chosen by the environment may either be influenced by a neighbour or relocate an incident edge to a different neighbour, based on \citet{holmeNonequilibriumPhaseTransition2006}. The environment is stochastic and fully observable.\\[0.2em]
\textit{Search and Rescue (SAR).} The agent must find and rescue as many survivors as possible by navigating a fragile network and transporting survivors to an exit node. Each node has a stability score, and movement through the node has a chance to reduce the stability, causing incident edges to collapse when below a threshold. The environment is stochastic and partially observable.\\[0.2em] %
\textit{Cascading Failures (CF).} Overloaded nodes in a network fail stochastically, causing incident edges to collapse and redistribute the load to neighbours \citep{motterCascadebasedAttacksComplex2002}. The agent accelerates this process by selecting a node to disconnect at each step. The safety margin of the nodes is determined by a random unobserved variable, making the environment stochastic and partially observable.

\paragraphsmall{Evaluation Procedure}
The primary evaluation metric is the mean per-episode GDD of the model's predictions under random rollouts, which we refer to as the GDD score.
Full evaluation details are provided in \Cref{app:evaluation_details}.
We note here that the GDD score is dependent on the environment's features and is not inherently size-invariant, so it is not directly comparable across environments or graph sizes.
However, it is a useful metric for evaluating the quality of a model's distributions, accounting for variance and capturing the dependence between node, graph, and topology features. 

\paragraphsmall{Training Details}
For each model and environment, we train for 5000 epochs on a set of pre-collected rollouts of a random policy on a set of graphs with sizes in $\{15, 18, 20, 23, 25\}$.
Hyperparameters were chosen from a grid search based on the mean GDD score achieved on a validation set consisting of 20 graphs of each size 20 and 30.
Full training details are provided in \Cref{app:training_details}.

\subsection{Size Generalisation for Next-State Distribution Predictions}

We evaluate the size generalisation of each model by calculating the single-step GDD score on test sets of different sizes, with 20 nodes being within the training distribution, 30 nodes being only in the validation set, and all larger sizes being completely outside the training distribution.
The results in \Cref{tab:ood_eval} demonstrate that the GDM outperforms or matches all baselines on the stochastic environments.
We do not evaluate RGC on large graphs due to episode lengths quadratic in $\numnodes$.

\begin{table}[tbh]
	\centering
	\caption{GDD score ($\downarrow$) for single-step predictions on test sets of different sizes (10 seeds).}\label{tab:ood_eval}
	\resizebox{0.95\linewidth}{!}{%
	\renewcommand{\arraystretch}{0.8}
	\input{tables/ood_eval.tex}	}
\end{table}

We validate the GDD score by analysing the per-factor distribution metrics in \Cref{app:factorwise_distribution_metrics}.
The GDD score is consistent with the trend of these metrics but is also sensitive to the joint distribution of all factors, which is not captured by factor-wise metrics.
GDM consistently demonstrates extremely low BCE on the adjacency matrix, owing to the explicit recurrent adjacency matrix.
Accurate topology prediction is arguably the key component of a graph-based world model, as feature predictions are meaningless without an understanding of the underlying network.

\subsection{Multi-Step Rollouts}
\begin{figure}[tbh]
	\centering
	\includegraphics[width=\linewidth]{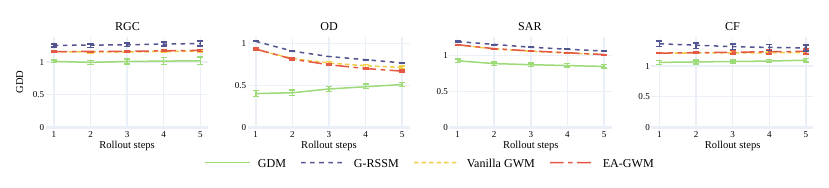}
	\caption{Multi-step rollout GDD on 50-node graphs ($\downarrow$).}
	\label{fig:multi-step-rollouts-main}
\end{figure}

We calculate the GDD score for rollouts of 2\textendash5 steps on 50-node graphs in \Cref{fig:multi-step-rollouts-main}, with full details in \Cref{app:multi-step-rollouts}.
GDM consistently outperforms the baselines.
Interestingly, there is little difference between Vanilla GWM and EA-GWM, suggesting that the regularisation terms in EA-GWM do not significantly improve multi-step rollout performance in these environments.

\subsection{Ablations}
We perform an ablation study to evaluate the contribution of each component of the GDM model.
We test the contribution of the separate graph state by removing $\hg(\graph)$ and $\zg(\graph)$. %
We test the effect of setting $\beta_h$ to $0$, which removes the loss term drawing $\vh^+$ and $\vh^-$ together.
We ablate the $\vh^+$ posterior correction by using only the prior $\vh^-$, aligning the model with the RSSM framework. %
We investigate the effect of the Hard Concrete sampling and SparseGAT message passing by using Bernoulli sampling for the adjacency update, and all-pairs message passing weighted by adjacency probabilities, respectively.
We also test the effect of removing the recurrent adjacency, instead using all-pairs message passing and only calculating $\appradj$ at the decoder output, as per G-RSSM.

\begin{figure}[t]
	\centering
	\includegraphics[width=\linewidth]{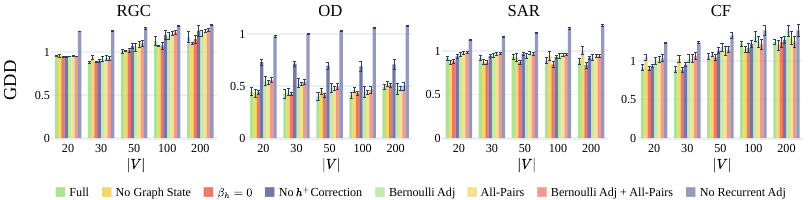}
	\caption{GDD scores for each ablation ($\downarrow$).}
	\label{fig:ablation}
\end{figure}

\Cref{fig:ablation} shows the GDD score at different test sizes for each ablation on the environments, with further results in \Cref{app:ablation_results}.
The removal of the graph component has variable effects across different environments.
Removing the $\beta_h$ loss term may slightly improve performance in particular environments.
The posterior correction has a significant effect in the OD environment but less in others, likely due to the random node feature in the OD environment which is difficult to encode in a sampled categorical latent variable.
The recurrent adjacency has a significant effect on the GDD across all environments.
The use of SparseGAT message passing and Hard Concrete sampling is more important in some environments than others.
Thus, we can conclude that the recurrent adjacency matrix is a key contributor to quality distribution predictions for evolving graphs.

%% file: tables/ood_eval.tex
\begin{tabular}{
 l
 l
 S[table-format=1.3, detect-weight] @{\scriptsize$\pm$} L
 S[table-format=1.3, detect-weight] @{\scriptsize$\pm$} L
 S[table-format=1.3, detect-weight] @{\scriptsize$\pm$} L
 S[table-format=1.3, detect-weight] @{\scriptsize$\pm$} L
 S[table-format=1.3, detect-weight] @{\scriptsize$\pm$} L
 S[table-format=1.3, detect-weight] @{\scriptsize$\pm$} L
 S[table-format=1.3, detect-weight] @{\scriptsize$\pm$} L
}
\toprule
 &  & \multicolumn{2}{c}{$\numnodes=20$} & \multicolumn{2}{c}{$\numnodes=30$} & \multicolumn{2}{c}{$\numnodes=50$} & \multicolumn{2}{c}{$\numnodes=100$} & \multicolumn{2}{c}{$\numnodes=200$} & \multicolumn{2}{c}{$\numnodes=500$} & \multicolumn{2}{c}{$\numnodes=1000$} \\
\midrule
RGC & GDM & \bfseries 0.950 & {\bfseries 0.007} & \bfseries 0.880 & {\bfseries 0.012} & \bfseries 1.010 & {\bfseries 0.021} & \bfseries 1.125 & {\bfseries 0.054} & 1.175 & 0.065 & \multicolumn{2}{c}{ -- } & \multicolumn{2}{c}{ -- } \\
RGC & Vanilla GWM & 1.168 & 0.002 & 1.158 & 0.003 & 1.156 & 0.004 & 1.161 & 0.005 & \bfseries 1.167 & {\bfseries 0.006} & \multicolumn{2}{c}{ -- } & \multicolumn{2}{c}{ -- } \\
RGC & EA-GWM & 1.167 & 0.002 & 1.161 & 0.005 & 1.160 & 0.008 & 1.170 & 0.007 & 1.177 & 0.006 & \multicolumn{2}{c}{ -- } & \multicolumn{2}{c}{ -- } \\
RGC & G-RSSM & 1.226 & 0.009 & 1.229 & 0.007 & 1.254 & 0.021 & 1.283 & 0.023 & 1.300 & 0.018 & \multicolumn{2}{c}{ -- } & \multicolumn{2}{c}{ -- } \\
 \midrule
OD & GDM & \bfseries 0.449 & {\bfseries 0.042} & \bfseries 0.426 & {\bfseries 0.041} & \bfseries 0.402 & {\bfseries 0.037} & \bfseries 0.413 & {\bfseries 0.027} & \bfseries 0.492 & {\bfseries 0.022} & \bfseries 0.650 & {\bfseries 0.018} & \bfseries 0.766 & {\bfseries 0.014} \\
OD & Vanilla GWM & 0.916 & 0.005 & 0.913 & 0.004 & 0.930 & 0.005 & 0.972 & 0.007 & 0.995 & 0.007 & 1.016 & 0.007 & 1.025 & 0.007 \\
OD & EA-GWM & 0.920 & 0.011 & 0.920 & 0.009 & 0.935 & 0.010 & 0.973 & 0.010 & 0.994 & 0.010 & 1.013 & 0.009 & 1.023 & 0.009 \\
OD & G-RSSM & 1.034 & 0.002 & 1.024 & 0.002 & 1.027 & 0.002 & 1.044 & 0.001 & 1.053 & 0.001 & 1.062 & 0.001 & 1.068 & 0.002 \\
\midrule
SAR & GDM & \bfseries 0.917 & {\bfseries 0.023} & \bfseries 0.917 & {\bfseries 0.028} & \bfseries 0.928 & {\bfseries 0.026} & \bfseries 0.887 & {\bfseries 0.032} & \bfseries 0.878 & {\bfseries 0.035} & \bfseries 0.886 & {\bfseries 0.038} & \bfseries 0.882 & {\bfseries 0.033} \\
SAR & Vanilla GWM & 1.122 & 0.004 & 1.131 & 0.003 & 1.150 & 0.003 & 1.180 & 0.007 & 1.202 & 0.008 & 1.224 & 0.007 & 1.227 & 0.006 \\
SAR & EA-GWM & 1.126 & 0.006 & 1.133 & 0.005 & 1.152 & 0.004 & 1.183 & 0.007 & 1.205 & 0.008 & 1.226 & 0.007 & 1.227 & 0.006 \\
SAR & G-RSSM & 1.139 & 0.005 & 1.164 & 0.005 & 1.196 & 0.005 & 1.232 & 0.005 & 1.256 & 0.005 & 1.286 & 0.015 & 1.302 & 0.025 \\
\midrule
CF & GDM & \bfseries 0.914 & {\bfseries 0.038} & \bfseries 0.890 & {\bfseries 0.043} & \bfseries 1.059 & {\bfseries 0.034} & \bfseries 1.222 & {\bfseries 0.033} & \bfseries 1.249 & {\bfseries0.034} & \bfseries 1.261 & {\bfseries 0.056} & \bfseries 1.261 & {\bfseries 0.063} \\
CF & Vanilla GWM & 1.179 & 0.002 & 1.198 & 0.003 & 1.212 & 0.004 & 1.227 & 0.006 & \bfseries 1.249 & {\bfseries 0.010} & 1.271 & 0.013 & 1.277 & 0.014 \\
CF & EA-GWM & 1.171 & 0.002 & 1.193 & 0.003 & 1.211 & 0.005 & 1.233 & 0.009 & 1.259 & 0.013 & 1.284 & 0.017 & 1.292 & 0.018 \\
CF & G-RSSM & 1.240 & 0.010 & 1.277 & 0.010 & 1.363 & 0.042 & 1.436 & 0.053 & 1.470 & 0.060 & 1.485 & 0.066 & 1.493 & 0.069 \\
\bottomrule
\end{tabular}

%% file: sections/limitations.tex
While the aim of the GDM model is to predict the next-state distribution, there is some mismatch between the training objective, which considers reconstruction error in expectation, and the evaluation metric, which accounts for the full variance.
Training the model to minimise a distributional score directly would require sampling multiple rollouts for each training transition.
A limitation of GDM, which is shared with the other approaches considered in this work, is the independent sampling of edges.
For some environments where edge events are inherently dependent, the model cannot accurately capture the dynamics, even if the probability of each edge is accurate. 
For the edge updates, GDM requires logit calculations for every pair of nodes, limiting the scalability.
The model also does not include edge features, which restricts the environments to which GDM can be applied.
Future work could explore the inclusion of edge features without $\mathcal{O}(\numnodes^2)$ computation.

A limitation of MMD is that it is kernel-dependent, reducing comparability across environments.
MMD also scales quadratically with the number of samples, which can be limiting for multi-step rollouts.
In addition, a characteristic kernel compresses the range of bad predictions, necessitating the addition of an extra term to the GDD metric to make it more informative, introducing an arbitrary weighting between the two terms.
Lastly, the GDD score represents the ability of a model to predict next-state distributions, but does not necessarily reflect the model's ability to support downstream tasks.
In future work we plan to explore the relationship between GDD and performance on downstream tasks, such as reinforcement learning, to better understand the utility of the metric.

%% file: sections/conclusion.tex
In this work, we have introduced the GDM, a graph-based world model for stochastic and partially observable environments with evolving topologies, and the GDD metric for evaluating the quality of predicted next-state distributions.
The GDM performs message passing on a sparse recurrent adjacency matrix and integrates a recurrent state-space model to capture stochastic transitions, enabling the modelling of stochastic dynamics in latent space with accurate topology predictions.
The GDD metric uses MMD with a characteristic graph kernel to compare predicted and true next-state distributions, capturing dependencies between topology, node features, and graph features, and provably distinguishing between different distributions.
We have demonstrated that the GDM outperforms the baselines in the GDD metric across a range of environments and graph sizes outside of the training distribution.
Future work could explore the inclusion of edge features, and the relationship between GDD and performance on downstream tasks such as reinforcement learning.

%% file: sections/appendix.tex
\section{Proofs}\label{app:proofs}

\subsection{Proof of \Cref{lemma:characteristic_joint_kernel}}\label{app:proof_characteristic_joint_kernel}
For the continuous features, after concatenating the coordinates over the aligned nodes and feature index sets, each term
$\exp\left(-\frac{\|x-x'\|^2}{2r^2}\right)$
is an RBF kernel under a positive diagonal rescaling on a finite-dimensional Euclidean space and is therefore $C_0$-universal.
Since $K_n^\continuous$ and $K_g^\continuous$ are finite positive mixtures of such kernels, they are also $C_0$-universal \citep{steinwart2001influence}.

For the categorical features, each feature-index set and each corresponding value space is finite.
If the categorical feature index set is empty, the distance is defined to be $0$, and the corresponding kernel is identically $1$ and can be omitted from the product.
For a finite set, a kernel is universal if and only if its Gram matrix is strictly positive definite \citep{borgwardtIntegratingStructuredBiological2006}. 
Let
\[
d_{ij}
=
\frac{1}{|\features_g^\categorical|}
\sum_{f\in\features_g^\categorical}
\lambda_f
\mathbf{1}\!\left[{\gf}_f(\graph_i)\neq {\gf}_f(\graph_j)\right].
\]
The corresponding Gram matrix is $\mK_{ij}=\exp(-d_{ij})$.
The distance $d$ is a weighted Hamming distance. 
Weighted Hamming distance is conditionally negative definite when the weights are nonnegative. 
Hence, by Schoenberg's theorem, $\mK$ is positive definite.
Moreover, because all weights satisfy $\lambda_f>0$, this kernel is the product of strictly positive definite kernels on the individual finite categorical coordinates, and is therefore strictly positive definite. 
Thus $K_g^\categorical$ is universal. The same argument applies to $K_n^\categorical$.

For the adjacency term, because the node set is finite, the set of possible edges is finite, and every graph topology is a subset of the finite set of possible edges.
\citet{bouchard2013proof} prove that the Jaccard kernel matrix on the non-empty subsets of any finite ground set is strictly positive definite.
Under the stated convention for the empty set, $K_J$ extends this matrix by an isolated diagonal entry $K_J(\emptyset,\emptyset)=1$, with $K_J(\emptyset,E)=0$ for $E\neq\emptyset$, and is therefore strictly positive definite on the full set of possible graph topologies.
Since this domain is finite, strict positive definiteness makes the Gram matrix invertible, so the RKHS can represent every real-valued function on the domain. 
Hence $K_J$ is universal on the finite space of graph topologies.

Finally, $K_\mathrm{joint}$ is the tensor-product kernel. 
The product of $C_0$-universal kernels is $C_0$-universal on the product domain. 
Therefore $K_\mathrm{joint}$ is $C_0$-universal and hence characteristic.
\hfill$\square$

\subsection{Proof of \Cref{lemma:characteristic_kernel}}\label{app:proof_characteristic_kernel}
Each continuous feature kernel $k_f^\continuous$ is a finite positive mixture of RBF kernels and is therefore positive definite. 
Each categorical feature kernel
$
k_f^\categorical(x,y)
=
\exp\left(-\lambda_f\mathbf{1}[x\ne y]\right)
$
is also positive definite: on the finite value space of feature $f$, its Gram matrix is 
$
(1-e^{-\lambda_f})I+e^{-\lambda_f}\mathbf{1}\mathbf{1}^\top,
$
which is positive semidefinite for $\lambda_f>0$. 
Hence $K_\mathrm{add}^n$ and $K_\mathrm{add}^g$, being nonnegative averages of positive-definite kernels, are positive definite.
As in the proof of \Cref{lemma:characteristic_joint_kernel}, $K_J$ is a positive-definite kernel on finite sets.
Therefore, since $w_n,w_g,w_e\geq0$,
$
K_\mathrm{add}
$
is positive definite.

By \Cref{lemma:characteristic_joint_kernel}, $K_\mathrm{joint}$ is characteristic. 
For any probability measures $P,Q$,
$
\operatorname{MMD}^2_K(P,Q)
=
\operatorname{MMD}^2_{K_\mathrm{add}}(P,Q)
+
\epsilon\operatorname{MMD}^2_{K_\mathrm{joint}}(P,Q),
$
where both terms on the right-hand side are nonnegative. 
Since $\epsilon>0$,
$
\operatorname{MMD}^2_K(P,Q)=0
\Rightarrow
\operatorname{MMD}^2_{K_\mathrm{joint}}(P,Q)=0.
$
Because $K_\mathrm{joint}$ is characteristic, this implies $P=Q$. Hence $K$ is characteristic.
\hfill$\square$

\section{Environment Descriptions}\label{app:environment_descriptions}

Each environment is implemented as a POMDP as described in \Cref{sec:mdp}.
Within each environment, the observation space consists of the adjacency matrix for the current graph, the categorical and continuous node features, and the categorical and continuous graph features.
These features are environment-specific and are described in detail below.
For completeness, we include the number of node selections $k$ in the graph features, and include the action mask in the node features.

\subsection{Search and Rescue}

The SAR environment is a stochastic, partially observable environment in which the agent navigates a graph to collect survivors and escort them to an exit node.

States consist of undirected graphs $\graph=(\nodes,\edges,\{\nf(v)\}_{v\in\nodes}, \gf(\graph))$.
The node features are
\[
\nf(v)=\{\mathrm{survivor\_present}(v),\mathrm{stability}(v),\mathrm{exit\_node}(v),\mathrm{current\_location}(v)\},
\]
where $\mathrm{survivor\_present}(v),\mathrm{exit\_node}(v),\mathrm{current\_location}(v)\in\{0,1\}$, and $\mathrm{stability}(v)\in[0,1]$.
The graph features are $\{\mathrm{escorting}, k\}$.
The $\mathrm{escorting}\in\{0,1\}$ feature indicates whether the agent is currently escorting a survivor.

The agent selects $k=1$ adjacent node at each transition and moves to that node.
If the agent is not escorting and the selected node contains a survivor, it collects the survivor.
If the agent is escorting and moves to the exit node, it delivers the survivor.
The environment begins with the agent at the exit node.

After the agent moves to node $u$, $u$ loses $0.3$ stability with probability $0.5$, unless $u$ is the exit node:
$
\mathrm{stability}(u)
\leftarrow
\max\{0,\mathrm{stability}(u)-0.3\}.
$
The environment then removes every edge $(u,v)$ incident to the visited node for which
$
\mathrm{stability}(u)+\mathrm{stability}(v)<1.
$
Thus, movement may disconnect parts of the graph over time.

The reward is $-1$ per transition, with an additional reward of $1$ for collecting a survivor, $10$ for delivering a survivor to the exit, and $5$ for rescuing all survivors.
An episode terminates successfully when all survivors have been delivered.
If the agent cannot reach the exit, the episode terminates with an additional penalty of
$
-30-5n_{\mathrm{remaining}},
$
where $n_{\mathrm{remaining}}$ is the number of undelivered survivors.
If the agent is not escorting a survivor and no remaining survivor is reachable from the exit, the episode terminates with an additional penalty of
$
-5n_{\mathrm{remaining}}.
$
The episode is truncated after a number of transitions equal to $\numnodes$ times the initial number of survivors.

To create the partial observation, the true survivor-presence feature is revealed only for nodes within two hops of the agent's current location; survivor locations elsewhere are reported as absent.
All other state features, including the graph topology, node stability, exit node, and current location, remain observable.

For current location $u$, the action mask is
$
\actionmask(s)
=
\{v\in\nodes\mid (u,v)\in\edges\}.
$
Thus, the agent must select exactly one adjacent node at every transition.

We sample the initial graph from a Delaunay triangulation of a set of $\numnodes$ points located in the unit square.
We select one exit node uniformly at random, select $30\%$ of nodes, excluding the exit, to have survivors, and sample initial node stabilities uniformly from $[0.5,1]$, setting the exit-node stability to $1$.
Parameters for the SAR environment were chosen via grid search to produce an environment with a reasonable margin between the performance of a greedy policy and a random policy without becoming trivial.

\subsection{Opinion Dynamics}
The OD environment is a stochastic, fully observable environment in which the agent selects a set of nodes to influence towards a target opinion.

States consist of undirected graphs $\graph=(\nodes,\edges,\{\mathrm{opinion}(v), \zeta(v)\}_{v\in\nodes}, \{\mathrm{k\_frac}, \mathrm{k\_frac}_{\mathrm{env}}\})$.
For the node features, $\mathrm{opinion}(v)\in\{0,\ldots,4\}$ is node $v$'s current opinion, and $\zeta(v) \in[-1,1]$ is its fixed latent ``vibe''. 
In the graph feature, $\mathrm{k\_frac}$ specifies the fraction of nodes selected by the agent and $\mathrm{k\_frac}_{\mathrm{env}}$ specifies the fraction of nodes selected by the environment.

We use dynamics based on the model of \citet{holmeNonequilibriumPhaseTransition2006}.
At each transition, the environment first selects $k_{\mathrm{env}}=\lceil \numnodes \mathrm{k\_frac}_{\mathrm{env}}\rceil$ nodes uniformly without replacement, with $\mathrm{k\_frac}_{\mathrm{env}}=0.1$.
For each selected node, it either rewires one incident edge with probability $\phi=0.458$~\citep{holmeNonequilibriumPhaseTransition2006}, or copies the opinion of one of its neighbours otherwise. 
Rewiring replaces an edge to a dissimilar-vibe neighbour with an edge to a same-opinion, similar-vibe non-neighbour. 
Let $d(u,v)=| \zeta(u)-\zeta(v) |$ denote the distance between two nodes' vibes, and let $T=0.1$ be the temperature.
When selecting a vibe-similar node from a candidate set $C$, the environment uses
\[
P(v\mid u,C)
=
\frac{\exp(-d(u,v)/T)}
{\sum_{w\in C}\exp(-d(u,w)/T)}.
\]
For a rewiring event, the existing neighbour to disconnect is instead selected with
\[
P(v\mid u,C)
=
\frac{\exp(+d(u,v)/T)}
{\sum_{w\in C}\exp(+d(u,w)/T)}.
\]
It then connects the node to a currently unconnected node with the same opinion, sampled using the first, vibe-similar distribution.
The agent then selects $k=\lceil \numnodes \mathrm{k\_frac}\rceil$ nodes, with $\mathrm{k\_frac}=0.05$.
Each selected node independently adopts the target opinion $0$ with probability $0.3$.
The reward is the increase in the fraction of nodes holding the target opinion.
Parameters for the OD environment were chosen via grid search to produce an environment with a reasonable margin between the performance of a greedy policy and a random policy, while still having some margin for improvement on the greedy policy.

With no intervention, the environment converges to a steady state within several hundred to several thousand transitions.
Thus we set the horizon to $\horizon=50$ to evaluate the model's ability to predict the short-term dynamics of the environment.

We sample the initial graph from a connected Erd\H{o}s--Rényi distribution with expected degree $4$, assign each node one of five opinions uniformly at random, and sample each node's vibe uniformly from $[-1,1]$. 
The action mask includes every node, so any distinct set of nodes is a valid action.

\subsection{Cascading Failures}
The CF environment is based on the model of \citet{motterCascadebasedAttacksComplex2002}, which describes a failure model of electrical power grids where each node has a load and a capacity, and overloaded nodes fail, redistributing their load to their neighbours.
We introduce stochasticity and partial observability to the model.

States consist of undirected graphs $\graph=(\nodes,\edges,\{\mathrm{load}(v),\mathrm{capacity}(v)\}_{v\in\nodes}, \{k, \xi\})$.
The ${\mathrm{load}(v) \in [0, 1]}$ describes the current load on node $v$, and $\mathrm{capacity}(v) \in [0, 1]$ describes the rated capacity of node $v$.
The actual capacity of each node is determined by the unobserved safety factor ${\xi\in[1.1, 1.4]}$, which is a single graph feature fixed for the entire episode.
Per \citet{motterCascadebasedAttacksComplex2002}, the load of each node is determined by the normalised betweenness centrality of the node, 
\[
c_t(v) = \frac{1}{P} \sum_{u, w \in V} \frac{\operatorname{paths}_t(u, w \mid v)}{\operatorname{paths}_t(u, w)},
\]
where $\operatorname{paths}_t(u, w)$ is the total number of shortest paths between nodes $u$ and $w$ at step $t$, $\operatorname{paths}_t(u, w \mid v)$ is the number of those paths that pass through node $v$, and $P$ is a normalisation factor given by the number of pairs of nodes in the graph.
The rated capacity of each node is fixed to $c_0(v)$, the load of the nodes in the network before any failures. 

The transition function is stochastic and proceeds as follows.
At step $t$, the agent selects $k=1$ node to disconnect from the graph, and all adjacent edges to that node are removed.
Additionally, we define stochastic failures: any node $v$ for which $\mathrm{load}_{t-1}(v) > \xi\mathrm{capacity}(v)$ fails according to 
\[
p(v\text{ failure}) = \min\left( 1, \frac{\mathrm{load}_{t-1}(v) - \xi\mathrm{capacity}(v)}{\xi\mathrm{capacity}(v)}\right).
\]
Failed nodes are disconnected from the graph by the removal of all incident edges. 
Subsequently, the loads of all nodes are recalculated based on the new graph topology.

Observations are given by $o_t = (\nodes_t, \edges_t, \{\mathrm{load}_t(v), \mathrm{capacity}(v)\}_{v\in\nodes_t}, \{k\})$.
The action space is the set of nodes which have at least one incident edge: $\actionmask(s) = \{v\in\nodes \mid \exists u\in\nodes, (u,v)\in\edges\}$.

The horizon is set to $\horizon = \numnodes/2$, and the episode terminates after $\horizon$ steps or when the graph is completely disconnected.
Each step incurs a reward of $-1/\horizon$, and there is a terminal penalty of $|\operatorname{lcc}(G_t)|/\numnodes$ if the horizon is reached before the graph is completely disconnected, where $\operatorname{lcc}(G_t)$ is the largest connected component of the graph at step $t$.

We sample the initial graph from a Schultz-Heitzig-Kurths distribution~\citep{schultzRandomGrowthModel2014}, with parameters $n_0=0.5\numnodes$, $p=0.5$, $q=0.5$, $r=1/3$, and $s=0.1$.

\subsection{Robust Graph Construction}
The RGC environment, introduced by \citet{darvariuGoaldirectedGraphConstruction2021}, is a deterministic, fully observable environment in which the agent selects $k=2$ nodes to connect in order to increase the robustness of the graph under node removal.
The states $\states$ consist of graphs $\graph=(\nodes,\edges, \emptyset,\{\tau, k\})$, where the graph feature $\tau$ represents the remaining edge addition budget as a fraction of the total number of potential edges in the graph ($\numnodes(\numnodes-1)/2$).
The transition adds an edge between the two selected nodes, and decreases the budget $\tau$ by $2/(\numnodes(\numnodes-1))$, unless an edge already exists between them.
The reward function is the expected increase in robustness of the graph under targeted node removal, where the robustness is measured by the fraction of nodes required to be removed in order to disconnect the graph, with nodes being selected in descending order of initial degree.
Termination occurs when the edge addition budget is exhausted, i.e., when $\tau=0$, or after $\horizon=\numnodes(\numnodes-1)/2$ steps.

We sample the initial graph from an Erd\H{o}s-R\'{e}nyi distribution with $p=0.2$, and set $\tau=5\%$.
For simplicity, we define $\actionmask(s)$ to be $\{v\in\nodes \mid \exists u\in\nodes, (u,v)\notin\edges\}$, which is the set of nodes with at least one available edge to add.
This means that the agent may select a pair of nodes which are already connected, resulting in a noop.

\section{Baseline Implementations}\label{app:baseline_adaptation}

We adapt the G-RSSM model to our environments by replacing its observation encoder and decoder with environment-specific networks that support the node- and graph-level scalar, categorical, and binary features defined by each environment.
Categorical inputs are one-hot encoded, and separate output heads predict each state feature using an appropriate distribution.
Per \citet{karacelebiLearningAdHoc2026}, graph-level features are broadcast to nodes during encoding and node-set actions are represented by a binary indicator at each node and supplied to the corresponding node-level recurrent transition.

The original adjacency decoder produces a fixed-size $\numnodes\times\numnodes$ output and therefore cannot be applied to graphs whose size differs from those used during training.
We replace it with a size-independent node-wise decoder.
For each node $v$, a shared MLP maps its deterministic and stochastic recurrent states to an embedding
$
\emb(v)=f_{\mathrm{dec}}(\hn(v),\zn(v))
$,
and the adjacency logit for a pair of nodes is
$
\ell_{uv}
=
\left(\mW_{\mathrm{adj}}\emb(u)\right)^\top
\left(\mW_{\mathrm{adj}}\emb(v)\right)
$.
Because the same functions are applied to every node and node pair, the number of decoder parameters is independent of $\numnodes$, allowing the adapted G-RSSM to operate on graphs of different sizes.
We also apply the feature scaling described in \Cref{app:feature_scaling} to the G-RSSM decoder outputs.

We similarly adapt EA-GWM and Vanilla GWM to multimodal, variable-size graph environments.
The original models operate on continuous observable node states, whereas our environments may contain scalar, categorical, and binary node and graph features.
We therefore introduce feature-specific encoders and prediction heads, represent categorical and binary predictions by their probabilities during differentiable rollouts, and retain node and graph features as distinct components of the predicted state.
Continuous features are predicted as residual changes to their current values as per the original formulation \citep{songUnderstandingRolloutError2026}.

The adapted EA-GWM and Vanilla GWM share the same architecture and differ only in their training objectives.
Vanilla GWM is trained using the supervised one-step losses for the next graph state, adjacency, reward, continuation, and action mask.
The EA-GWM additionally uses rollout-consistency, spectral, and critical-node losses.
We generalise the rollout and critical-node losses to use mean-squared error for scalar factors and classification losses for categorical and binary factors.
Critical-node weighting is applied only to node-level state losses, with node weights proportional to degree and normalised to have mean one within each graph.

\paragraphsmall{Adjacency Sampling}
The GDM model uses Hard Concrete sampling during imagination to produce a sparse adjacency matrix update, applied to the previous recurrent adjacency matrix.
The decoder for GDM is a pass-through network that outputs the same predicted adjacency matrix without resampling.
For EA-GWM and Vanilla GWM, the adjacency matrix is predicted by a learned transformation of the node features: ${\appradj(i,j) = \sigma(\vh(i)^\top \mQ \vh(j))}$.
Edges are sampled from the Bernoulli distribution defined by $\appradj(i,j)$, and $\mQ$ is regularised to have a norm of $1$.
Due to the Bernoulli sampling, the adjacency matrix distribution can be soft, even though the model is deterministic.
This is disadvantageous in environments where transitions are stochastic but sharp, such as the SAR environment.
For the G-RSSM model, the adjacency matrix is predicted only by a decoder head and not used in the recurrent transition.

\section{Evaluation Details}\label{app:evaluation_details}
The GDD score is calculated from predictions of the model under random rollouts.
We use fixed test sets of 20 graphs of each size $\numnodes\in\{20, 30, 50, 100, 200, 500, 1000\}$, except for the RGC environment where the largest size is 200 due to episode length exceeding 6000 steps at 500 nodes.
For each graph, we perform a random rollout of an episode to obtain a set of reference states.
For a single transition, the environment is reset to the reference state and advanced one step using the reference action $\numsamples=30$ times to create the environment's transition distribution.
Similarly, the model's transition distribution is obtained by performing a forward pass from the reference state and action $\numsamples$ times.
For latent-state models, the model's initial latent state is based on the posterior-corrected latent state from the preceding trajectory, while for state-space models, the model's state is based on the environment's reference state.

The transition GDD is calculated using \Cref{eqn:GDD} with all pairs of samples from the environment and model transition distributions, and the mean per-episode GDD is calculated by averaging the transition GDD across transitions in the episode.
In some environments, the episode length varies based on the size of the graph, so for computational efficiency, we select 20 reference transitions per episode, always including the first and final transitions.
For continuous and categorical features, bandwidths are fitted based on the training set using the median heuristic, with full details provided in \Cref{app:bandwidth_fitting}.

In the calculation of GDD, we include the reward, continuation, and action mask in the state features, so that the GDD score reflects the model's ability to predict these factors as well as the next graph state.
The reward is considered a continuous graph feature, the continuation is a categorical graph feature, and the action mask is a categorical node feature.
In a terminal transition, we consider only the reward and continuation, as the action mask and observation are not defined after termination.

\subsection{Bandwidth Fitting}\label{app:bandwidth_fitting}

We fit MMD bandwidths for categorical and continuous node features separately for each environment, using transitions sampled from the training rollouts.
For each environment we collect $8$ transitions from each episode in the training set, resulting in a total of $8\times100\times5=4000$ transitions for each environment.
All fitted continuous scales and categorical weights are precomputed and reused for every model evaluated in that environment.

For each continuous node or graph feature $f$, we compute the absolute one-step changes
\[
\Delta_f
=
\left\{
\left|x_{t+1,f}-x_{t,f}\right|
:
\left|x_{t+1,f}-x_{t,f}\right|>0
\right\}
\]
over all collected transitions.
For node features, changes are pooled over all nodes.
The base bandwidth is the median nonzero one-step change,
\[
b_f=\operatorname{median}(\Delta_f).
\]
If a feature never changes in the sampled transitions, we instead use the median nonzero absolute value observed for that feature.
If the feature is identically zero, we use the minimum bandwidth $10^{-8}$.
The multi-bandwidth RBF kernel uses bandwidth multipliers $
\mathcal{M}=\{0.1,0.5,1,2,16\}
$.

For categorical features, we fit a weight from the empirical one-step change rate.
For every transition, we compute the fraction $\rho_{t,f}$ of valid entries whose value changes and set
\[
\lambda_f
=
\frac{1}{\mathbb{E}_t[\rho_{t,f}]}.
\]
Thus, changes in normally stable categorical features receive greater weight than changes in frequently varying features.
If no changes are observed, we set $\lambda_f=1$.

\section{Experiment Details}

\subsection{Feature Scaling}\label{app:feature_scaling}

To prevent features with large magnitudes from dominating the decoder objective, we standardise each scalar prediction target using statistics calculated from the training data.
For each scalar feature $f$, we calculate the mean $\mu_f$ and population standard deviation $\sigma_f$.
The statistics are calculated over the scalar values in all stored pre- and post-transition observations.
The reward mean and standard deviation are fitted separately using all transition rewards.
If a target is constant throughout the training data, we set $\sigma_f=1$.

Each scalar decoder head produces a value $\hat z_f$ in standardised coordinates.
Before exposing the prediction to the model or evaluator, we transform it back to the original environment units:
$
\hat x_f
=
\sigma_f\hat z_f+\mu_f.
$
The reconstruction loss is calculated in standardised coordinates:
\[
\mathcal{L}_f
=
\left(
\frac{\hat x_f-\mu_f}{\sigma_f}
-
\frac{x_f-\mu_f}{\sigma_f}
\right)^2
=
\left(
\frac{\hat x_f-x_f}{\sigma_f}
\right)^2.
\]
Consequently, decoder predictions and sampled trajectories remain expressed in the original units of the environment, while each scalar feature contributes error relative to its variation in the training data.
The same procedure is applied to the reward head.

\subsection{Training Details}\label{app:training_details}

For each model, we train for 5000 epochs on a set of pre-collected rollouts of a random policy on a set of graphs with sizes in $\{15, 18, 20, 23, 25\}$ (100 graphs per size).
The random policy uniformly samples from the valid actions at each transition, so the training data never contains an invalid action.
We use the Adam optimiser with a batch size of 8 and a sequence length of 20, training 10 seeds for each model.
Hyperparameters for each model/environment pair were chosen from a grid search based on the mean GDD score achieved on a validation set consisting of 20 graphs of each node count 20 and 30.
Hyperparameter details are provided in \Cref{app:hyperparameter_search}.

\subsection{Hyperparameter Search}\label{app:hyperparameter_search}
The parameter search ranges for each model are provided in \Cref{tab:hyperparameter_search}.
Each combination was trained with 10 different seeds, and the minimum mean GDD score on the validation set over the seeds was used to select the best hyperparameters for each model/environment pair.
The $L$ parameter represents the number of rounds of message passing, being GNN layers for GDM and GWM and rounds of recurrent message passing for G-RSSM.
For G-RSSM the Encoder $L$ represents the number of GNN layers in the encoder.
\begin{table}[th]
	\centering
	\caption{Hyperparameter search ranges for each model.}\label{tab:hyperparameter_search}
	\resizebox{\linewidth}{!}{%
	\input{tables/hparam_search.tex}
	}
\end{table}

\subsubsection{Final Hyperparameters}
The final hyperparameters for each model and environment are provided in \Cref{tab:final_hyperparameters}.
Across environments, we use Hard Concrete parameters of $\beta=2/3$, $\gamma=-0.1$, and $\zeta=1.1$.
We use the pooling function $\pool=\operatorname{mean}$ for all models and environments.

\begin{table}[th]
	\centering
	\caption{Final hyperparameters for each model and environment. Unsearched parameters are indicated with an asterisk (*).}\label{tab:final_hyperparameters}
	\input{tables/hparam_final.tex}
\end{table}

\clearpage

\section{Additional Results}\label{app:additional_results}

\subsection{GDD Verification}\label{app:zero-GDD}

For the calculation of GDD, we use the biased MMD estimator in \Cref{eqn:mmd_estimate}, which may produce small non-zero values between identical distributions due to finite sample effects.
To measure the magnitude of this effect, we calculate the GDD between the reference environment distribution and itself under the same sampling procedure used for model evaluation. 
The results are provided in \Cref{tab:self_mmd}, which shows that the GDD bias is negligible for all environments and graph sizes.
Results for RGC are exactly zero because the environment is deterministic, and therefore the reference distribution is a single point mass.

\begin{table}[th]
	\centering
	\caption{Self-GDD between the reference environment distribution and a sampled environment distribution, with $\numsamples = 30$ samples.}
	\label{tab:self_mmd}
	\input{tables/self_mmd.tex}
\end{table}

\subsection{Training Curves}

\begin{figure}[th]
	\centering
	\includegraphics[width=\linewidth]{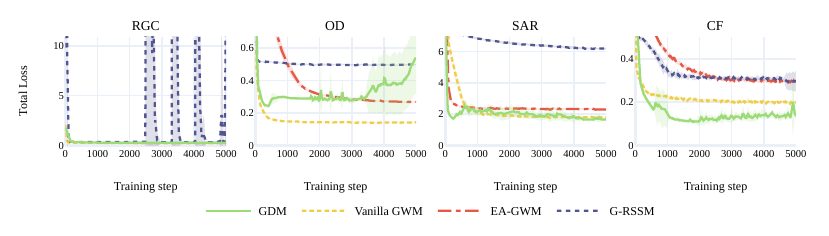}
	\caption{Training loss curves for each method/environment, averaged across 10 seeds.}\label{fig:loss_curve}
\end{figure}

\Cref{fig:loss_curve} shows the total loss curves, smoothed over 10 points, for each environment during training.
Note that the loss equation and coefficients can differ between methods, so the values are not directly comparable across methods.
In the OD environment, the loss for GDM increases after an initial stable period, which is due to the model learning to predict the reward function at the expense of higher KL loss in the node features.

\subsection{Factor-Wise Distribution Metrics}\label{app:factorwise_distribution_metrics}

\begin{figure}[th]
	\centering
	\includegraphics[width=\linewidth]{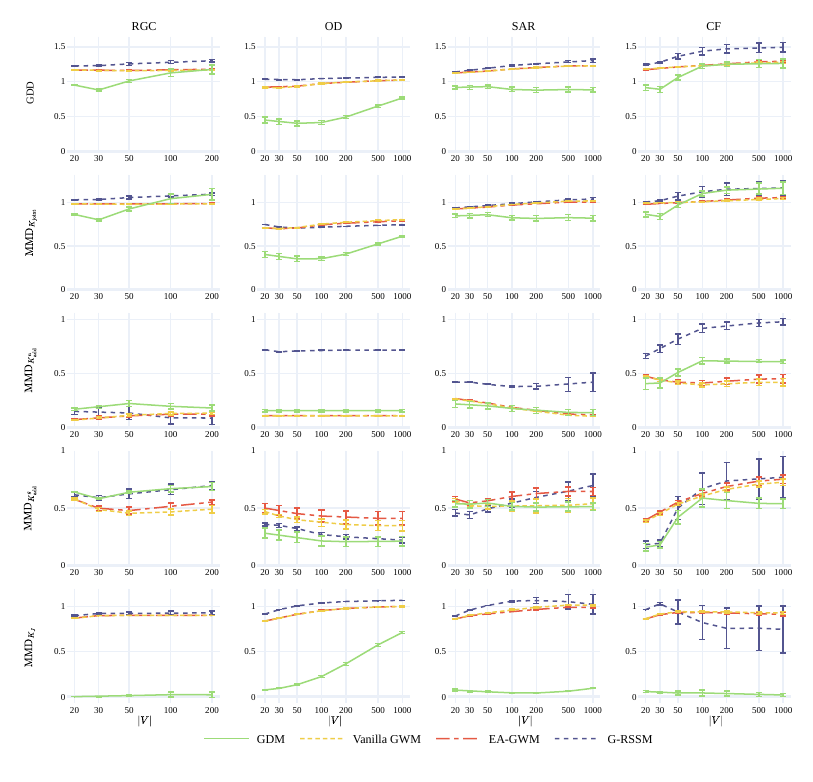}
	\caption{GDD score and individual components of the GDD for each environment and model.}
	\label{fig:gdd-components}
\end{figure}

\begin{figure}[th]
	\centering
	\includegraphics[width=\linewidth]{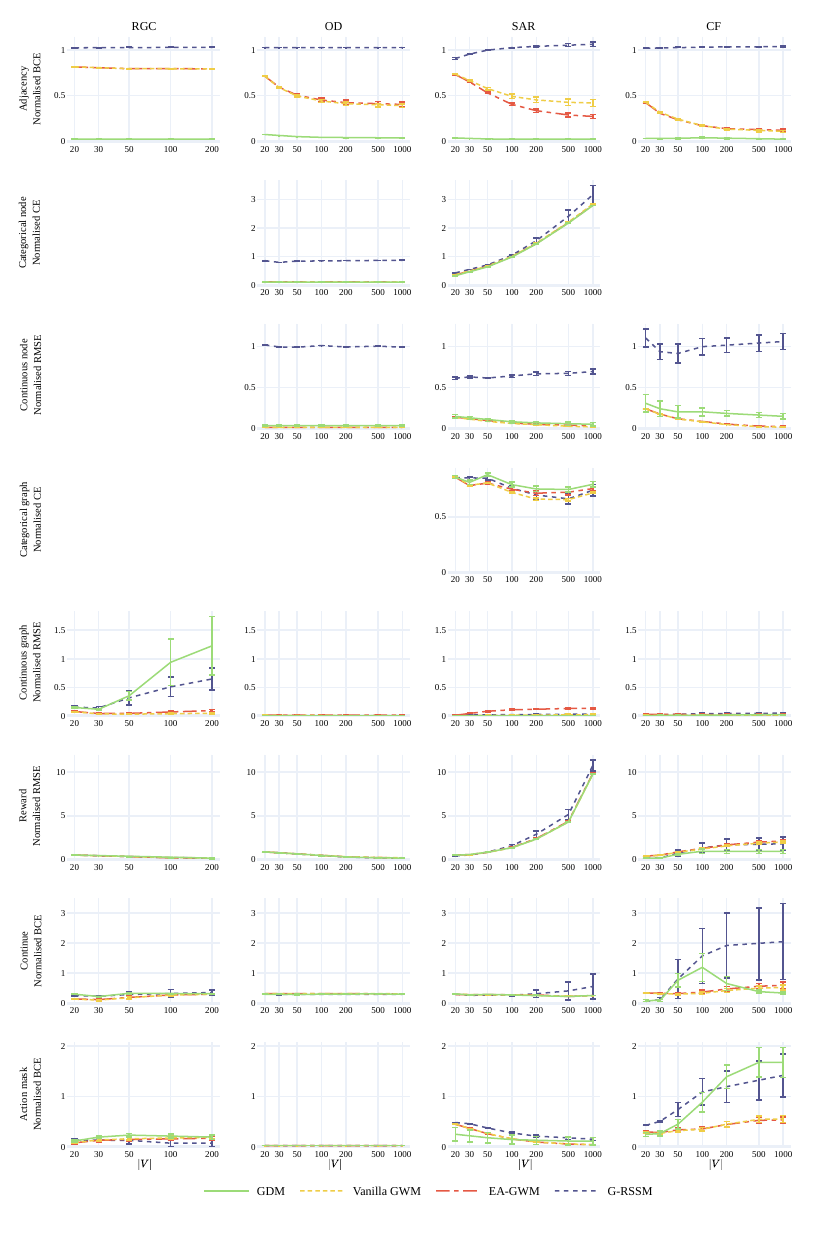}
	\caption{Factor-wise distribution metrics for each environment. Empty plots indicate absence of features for that environment.}
	\label{fig:factor-metrics}
\end{figure}

For single-step predictions, we show the breakdown of the four components of GDD score in \Cref{fig:gdd-components}.
We also show the full factor-wise distribution metrics for each environment in \Cref{fig:factor-metrics}.
GDM consistently predicts the adjacency matrix with extremely low BCE, demonstrating that the model is able to accurately capture the graph topology.
The G-RSSM model does not explicitly model the adjacency matrix, and therefore has high BCE for the adjacency in all environments.
The Vanilla GWM and EA-GWM models are able to make some predictions for the adjacency matrix, but struggle especially with static edges.
Since the adjacency matrix is predicted from node features and there is no conditioning on previous adjacency, these models struggle to predict evolving topologies where many edges remain unchanged across transitions.

The Vanilla GWM and EA-GWM models are able to capture certain features more accurately than the G-RSSM and GDM models. 
This is due to a limitation of the RSSM framework, which requires the model to incorporate all posterior information through the sampled stochastic latent state.
This means that certain information, particularly information with high variance across episodes and little redundancy, can be difficult to represent accurately in the latent state.
The GDM model partially mitigates this issue by using a deterministic state update on the posterior path, which improves its representational ability relative to G-RSSM but can still struggle with information loss.

\clearpage

\subsection{Multi-Step Rollout Results}\label{app:multi-step-rollouts}

\begin{figure}[th]
	\centering
	\includegraphics[width=\linewidth]{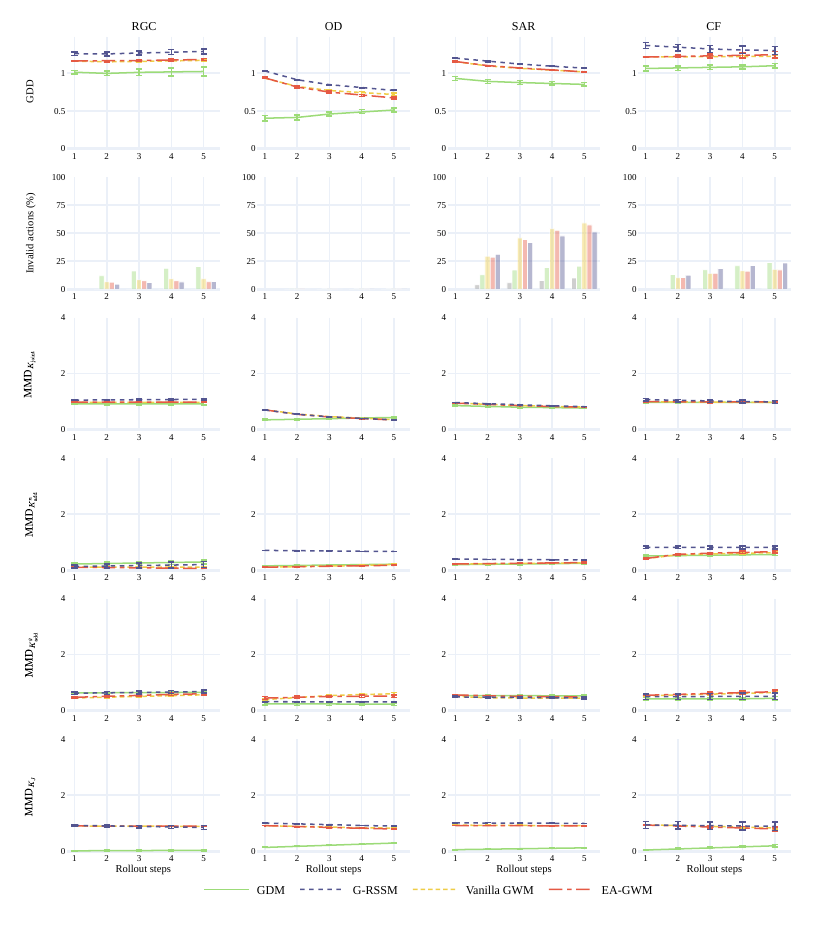}
	\caption{Multi-step rollout GDD, invalid action rate, and MMD components on 50-node graphs.}
	\label{fig:multi-step-rollouts}
\end{figure}

When calculating the GDD score for multi-step rollouts, we select initial states as per single-step rollouts, then perform a rollouts of $T$ steps according to the pre-determined actions.
In some cases, this can lead to invalid actions being taken, as the stochastic transition may result in a state where the action mask does not permit the action that was valid in the reference trajectory.
For the purpose of the rollouts, we allow invalid actions in the environment by performing a noop, and in the environment by running the forward pass with the invalid action.
Due to the diffusion of the distribution with more steps, we sample 100 points for the multi-step GDD calculation.
\Cref{fig:multi-step-rollouts} shows the breakdown of the GDD score for each environment and model across rollout lengths from 1 to 5 steps, as well as the rate of invalid actions taken during the rollout, with the environment shown in grey.

\clearpage

\subsection{Ablation Results}\label{app:ablation_results}
The GDD for all ablation studies is found in \Cref{tab:ablation_full}.
We additionally show an ablation combining the removal of graph state, $\vh^+$ correction, and recurrent adjacency.

\begin{table}[th]
	\centering
	\caption{GDD score (single-step) for each ablation and environment. Mean and 95\% confidence interval over 10 seeds.}\label{tab:ablation_full}
	\resizebox{\linewidth}{!}{%
	\input{tables/ablation_full.tex}
	}
\end{table}

%% file: tables/hparam_search.tex
\begin{tabular}{llll}
	\toprule
	{Parameter} & {GDM} & {EA-/Vanilla GWM} & {G-RSSM} \\
	\midrule
	LR & $10^{-3}, 5\times 10^{-4}, 10^{-4}$ & $10^{-3}, 5\times 10^{-4}, 10^{-4}$ & $5\times10^{-3}, 10^{-3}, 5\times 10^{-4}, 10^{-4}$ \\
	$L$ & $1,2,3,4$ & $1,2,3,4$ & $1,2,3$ \\
	Capacity &\begin{tabular}{@{}c@{}}$\dim(\vh)=64,\, \dim(\zn)=32\times32$ \\ $\dim(\vh)=128,\, \dim(\zn)=16\times16$\end{tabular} & $\dim(\vh)=64, 128$ & $\dim(\vh)=128$\\
	$\beta_\text{adj}$ & $1$ & $0.1,1,10$ & $0.1,1,10$ \\
	Encoder $L$ & - & - & $1, 2, 3$ \\
	$\beta_\text{rep}$ & $0.1, 0.05, 0.01$ & - & $0.1$ \\
	\bottomrule
\end{tabular}

%% file: tables/hparam_final.tex
\begin{tabular}[t]{lllll}
	\toprule
	{Parameter} & {RGC} & {OD} & {SAR} & {CF} \\
	\midrule
	\multicolumn{5}{c}{\textit{GDM}}\\
	\midrule
	LR & $10^{-4}$ & $5\times10^{-4}$ & $5\times10^{-4}$ & $5\times10^{-4}$ \\
	$L$ & 1 & 1 & 1 & 1 \\
	$\dim(e)$ & 64 & 128 & 128 & 128 \\
	$\dim(\vh)$ & 64 & 128 & 128 & 128 \\
	$\dim(\zn(\nodes))$ & 32$\times$32 & 16$\times$16 & 16$\times$16 & 16$\times$16 \\
	$\dim(\zg(\graph))$* & 8$\times$8 & 8$\times$8 & 8$\times$8 & 8$\times$8 \\
	$\beta_\text{adj}$* & 1 & 1 & 1 & 1 \\
	$\beta_\text{dyn}$* & 1 & 1 & 1 & 1 \\
	$\beta_\text{rep}$ & 0.1 & 0.05 & 0.1 & 0.1 \\
	$\beta_\text{pred}$* & 1 & 1 & 1 & 1 \\
	$\beta_n$* & 1 & 1 & 1 & 1 \\
	$\beta_g$* & 1 & 1 & 1 & 1 \\
	$\beta_h$* & 1 & 1 & 1 & 1 \\
	$\freebits$* & 0 & 0.01 & 0.01 & 0.01 \\
	\midrule
	\multicolumn{5}{c}{\textit{EA-GWM}}\\
	\midrule
	LR & $10^{-3}$ & $10^{-4}$ & $5\times10^{-4}$ & $10^{-4}$ \\
	$L$ & 2 & 4 & 2 & 4 \\
	$\dim(\vh)$ & 64 & 64 & 128 & 128 \\
	$\beta_\text{adj}$ & 1 & 1 & 1 & 1 \\
	\midrule
	\multicolumn{5}{c}{\textit{Vanilla GWM}}\\
	\midrule
	LR & $5\times10^{-4}$ & $10^{-4}$ & $10^{-4}$ & $10^{-4}$  \\
	$L$ & 2 & 4 & 2 & 1 \\
	$\dim(\vh)$ & 64 & 128 & 128 & 128 \\
	$\beta_\text{adj}$ & 0.1 & 0.1 & 0.1 & 0.1 \\
	\midrule
	\multicolumn{5}{c}{\textit{G-RSSM}}\\
	\midrule
	LR & $10^{-3}$ & $10^{-3}$ & $10^{-3}$ & $10^{-3}$\\
	$L$ & 3 & 1 & 1 & 1 \\
	Encoder $L$ & 1 & 1 & 2 & 1 \\
	$\dim(e)$* & 128 & 128 & 128 & 128 \\
	$\dim(\vh)$* & 128 & 128 & 128 & 128 \\
	$\dim(\vz)$* & 8$\times$8 & 8$\times$8 & 8$\times$8 & 8$\times$8 \\
	$\beta_\text{adj}$ & 10 & 1 & 10 & 0.1 \\
	$\beta_\text{dyn}$* & 1 & 1 & 1 & 1 \\
	$\beta_\text{rep}$* & 0.1 & 0.1 & 0.1 & 0.1 \\
	$\beta_\text{pred}$* & 1 & 1 & 1 & 1 \\
	$\freebits$* & 0 & 0.01 & 0.01 & 0.01 \\
	\bottomrule
\end{tabular}

%% file: tables/self_mmd.tex
\begin{tabular}{
 l
 S[table-format=1.3, detect-weight] @{\scriptsize$\pm$} L
 S[table-format=1.3, detect-weight] @{\scriptsize$\pm$} L
 S[table-format=1.3, detect-weight] @{\scriptsize$\pm$} L
 S[table-format=1.3, detect-weight] @{\scriptsize$\pm$} L
 S[table-format=1.3, detect-weight] @{\scriptsize$\pm$} L
}
\toprule
Env. & \multicolumn{2}{c}{$\numnodes=20$} & \multicolumn{2}{c}{$\numnodes=30$} & \multicolumn{2}{c}{$\numnodes=50$} & \multicolumn{2}{c}{$\numnodes=100$} & \multicolumn{2}{c}{$\numnodes=200$} \\
\midrule
RGC & 0.000 & 0.000 & 0.000 & 0.000 & 0.000 & 0.000 & 0.000 & 0.000 & 0.000 & 0.000 \\
OD & 0.031 & 0.014 & 0.033 & 0.012 & 0.036 & 0.007 & 0.035 & 0.006 & 0.035 & 0.004 \\
SAR & 0.012 & 0.014 & 0.012 & 0.012 & 0.010 & 0.011 & 0.007 & 0.008 & 0.006 & 0.006 \\
CF & 0.003 & 0.011 & 0.002 & 0.010 & 0.003 & 0.010 & 0.002 & 0.009 & 0.002 & 0.007 \\
\bottomrule
\end{tabular}

%% file: tables/ablation_full.tex
\begin{tabular}{
 l
 S[table-format=1.3, detect-weight] @{\scriptsize$\pm$} L
 S[table-format=1.3, detect-weight] @{\scriptsize$\pm$} L
 S[table-format=1.3, detect-weight] @{\scriptsize$\pm$} L
 S[table-format=1.3, detect-weight] @{\scriptsize$\pm$} L
 S[table-format=1.3, detect-weight] @{\scriptsize$\pm$} L
}
\toprule
Ablations & \multicolumn{2}{c}{$\numnodes=20$} & \multicolumn{2}{c}{$\numnodes=30$} & \multicolumn{2}{c}{$\numnodes=50$} & \multicolumn{2}{c}{$\numnodes=100$} & \multicolumn{2}{c}{$\numnodes=200$} \\
\midrule
\multicolumn{11}{c}{\textit{RGC}} \\
\midrule
Full & 0.950 & 0.007 &  0.880 & { 0.012} & 1.010 & 0.021 & 1.125 & 0.054 & 1.175 & 0.065 \\
No Graph State & 0.955 & 0.013 & 0.933 & 0.023 &  1.009 & { 0.004} &  1.069 & { 0.009} &  1.100 & { 0.016} \\
$\beta_h = 0$ &  0.944 & { 0.005} & 0.886 & 0.006 & 1.021 & 0.022 & 1.070 & 0.037 & 1.148 & 0.048 \\
No $\vh^+$ Correction & 0.947 & 0.004 & 0.895 & 0.013 & 1.070 & 0.032 & 1.197 & 0.053 & 1.246 & 0.061 \\
Bernoulli Adj & 0.951 & 0.003 & 0.920 & 0.026 & 1.053 & 0.042 & 1.181 & 0.040 & 1.215 & 0.034 \\
All-Pairs & 0.952 & 0.006 & 0.931 & 0.033 & 1.089 & 0.033 & 1.214 & 0.023 & 1.242 & 0.018 \\
Bernoulli Adj + All-Pairs & 0.950 & 0.003 & 0.926 & 0.027 & 1.103 & 0.032 & 1.228 & 0.020 & 1.253 & 0.015 \\
No Recurrent Adj & 1.238 & 0.001 & 1.242 & 0.007 & 1.275 & 0.008 & 1.303 & 0.006 & 1.310 & 0.005 \\
No $h^+$ Correction, No Graph State, No Recurrent Adj & 1.238 & 0.001 & 1.245 & 0.002 & 1.363 & 0.023 & 1.496 & 0.046 & 1.542 & 0.048 \\
\midrule
\multicolumn{11}{c}{\textit{OD}} \\
\midrule
Full & 0.449 & 0.042 & 0.426 & 0.041 &  0.402 & { 0.037} &  0.413 & { 0.027} & 0.492 & 0.022 \\
No Graph State &  0.428 & { 0.037} & 0.441 & 0.034 & 0.450 & 0.029 & 0.465 & 0.027 & 0.523 & 0.024 \\
$\beta_h = 0$ & 0.441 & 0.019 &  0.425 & { 0.018} & 0.413 & 0.019 & 0.432 & 0.019 & 0.508 & 0.019 \\
No $\vh^+$ Correction & 0.730 & 0.028 & 0.716 & 0.024 & 0.696 & 0.033 & 0.690 & 0.046 & 0.708 & 0.044 \\
Bernoulli Adj & 0.549 & 0.043 & 0.529 & 0.040 & 0.485 & 0.045 & 0.446 & 0.052 &  0.474 & { 0.052} \\
All-Pairs & 0.539 & 0.019 & 0.522 & 0.018 & 0.476 & 0.018 & 0.439 & 0.020 & 0.477 & 0.019 \\
Bernoulli Adj + All-Pairs & 0.559 & 0.027 & 0.538 & 0.025 & 0.498 & 0.030 & 0.467 & 0.034 & 0.499 & 0.037 \\
No Recurrent Adj & 0.977 & 0.006 & 1.001 & 0.005 & 1.028 & 0.005 & 1.062 & 0.005 & 1.081 & 0.004 \\
No $h^+$ Correction, No Graph State, No Recurrent Adj & 0.980 & 0.007 & 1.004 & 0.011 & 1.031 & 0.017 & 1.070 & 0.024 & 1.089 & 0.026 \\
\midrule
\multicolumn{11}{c}{\textit{SAR}} \\
\midrule
Full & 0.917 & 0.023 & 0.917 & 0.028 & 0.928 & 0.026 & 0.887 & 0.032 & 0.878 & 0.035 \\
No Graph State &  0.865 & { 0.026} & 0.872 & 0.027 & 0.925 & 0.043 & 0.941 & 0.046 & 1.001 & 0.045 \\
$\beta_h = 0$ & 0.877 & 0.020 &  0.865 & { 0.024} &  0.871 & { 0.024} &  0.844 & { 0.030} &  0.835 & { 0.032} \\
No $\vh^+$ Correction & 0.937 & 0.020 & 0.938 & 0.016 & 0.962 & 0.015 & 0.927 & 0.019 & 0.920 & 0.018 \\
Bernoulli Adj & 0.958 & 0.023 & 0.948 & 0.025 & 0.942 & 0.026 & 0.936 & 0.029 & 0.920 & 0.033 \\
All-Pairs & 0.977 & 0.015 & 0.966 & 0.014 & 0.972 & 0.016 & 0.955 & 0.017 & 0.946 & 0.018 \\
Bernoulli Adj + All-Pairs & 0.982 & 0.010 & 0.971 & 0.012 & 0.966 & 0.013 & 0.957 & 0.014 & 0.940 & 0.017 \\
No Recurrent Adj & 1.123 & 0.007 & 1.159 & 0.007 & 1.207 & 0.007 & 1.256 & 0.008 & 1.288 & 0.009 \\
No $h^+$ Correction, No Graph State, No Recurrent Adj & 1.065 & 0.009 & 1.124 & 0.005 & 1.201 & 0.005 & 1.264 & 0.010 & 1.307 & 0.019 \\
\midrule
\multicolumn{11}{c}{\textit{CF}} \\
\midrule
Full & 0.914 & 0.038 & 0.890 & 0.043 & 1.059 & 0.034 & 1.222 & 0.033 & 1.249 & 0.034 \\
No Graph State & 1.049 & 0.039 & 1.028 & 0.041 & 1.089 & 0.027 &  1.152 & { 0.036} &  1.208 & { 0.064} \\
$\beta_h = 0$ &  0.903 & { 0.024} &  0.885 & { 0.033} &  1.045 & { 0.042} & 1.171 & 0.053 & 1.244 & 0.061 \\
No $\vh^+$ Correction & 0.932 & 0.019 & 0.952 & 0.028 & 1.133 & 0.039 & 1.225 & 0.046 & 1.276 & 0.044 \\
Bernoulli Adj & 1.002 & 0.042 & 1.029 & 0.051 & 1.180 & 0.053 & 1.317 & 0.063 & 1.396 & 0.072 \\
All-Pairs & 1.018 & 0.041 & 1.030 & 0.049 & 1.136 & 0.057 & 1.244 & 0.072 & 1.301 & 0.086 \\
Bernoulli Adj + All-Pairs & 1.035 & 0.046 & 1.062 & 0.046 & 1.150 & 0.044 & 1.216 & 0.063 & 1.248 & 0.074 \\
No Recurrent Adj & 1.239 & 0.006 & 1.239 & 0.011 & 1.332 & 0.033 & 1.396 & 0.060 & 1.396 & 0.075 \\
No $h^+$ Correction, No Graph State, No Recurrent Adj & 1.200 & 0.015 & 1.224 & 0.008 & 1.286 & 0.011 & 1.351 & 0.025 & 1.386 & 0.042 \\
\bottomrule
\end{tabular}